\documentclass[10pt,journal,compsoc]{IEEEtran}

\newif\ifsubmission\submissionfalse
\newif\ifdraft\draftfalse
\newif\ifanon\anonfalse

\usepackage[hyphens]{url}  
\usepackage{graphicx} 
\usepackage{xcolor}
\usepackage{colortbl}
\usepackage{lscape}
\usepackage{amssymb}
\usepackage{framed}
\usepackage{xspace}
\usepackage{booktabs}
\usepackage{xltabular}
\usepackage{supertabular}
\usepackage[T2A,T1]{fontenc}
\usepackage[utf8]{inputenc}
\usepackage[bulgarian,english]{babel}
\usepackage[numbers,sort]{natbib} 
\definecolor{doctestcolor}{RGB}{230, 239, 255}
\definecolor{errorcolor}{RGB}{255,232,236}
\definecolor{signaturecolor}{RGB}{252, 232, 235}

\definecolor{Gray}{gray}{0.85}
\newcolumntype{g}{>{\columncolor{Gray}}l}
\newcolumntype{C}{>{\centering\arraybackslash}X}

\definecolor{wellesleyblue}{RGB}{0, 39, 118}
\definecolor{oberlinred}{RGB}{207,16,45}
\definecolor{cornellred}{RGB}{255,0,0}
\newcommand\btrule[1]{\specialrule{#1}{0pt}{0pt}}
\newcommand{\niche}{\textsc{Niche}\xspace}
\newcommand{\med}{\textsc{Medium}\xspace}
\newcommand{\low}{\textsc{Low}\xspace}
\newcommand{\high}{\textsc{High}\xspace}
\newcommand{\nlang}{18\xspace}

\newcommand{\system}{MultiPL-E\xspace}
\newcommand{\newHE}{MultiPL-HumanEval\xspace}
\newcommand{\newMBPP}{MultiPL-MBPP\xspace}

\usepackage{float}
\usepackage{newfloat}
\usepackage{listings}

\floatstyle{ruled}
\newfloat{listing}{tb}{lst}{}
\floatname{listing}{Listing}

\usepackage{pervasives}

\renewcommand{\citet}[1]{\cite{#1}}
\renewcommand{\citep}[1]{\cite{#1}}

\title{\system: A Scalable and Extensible Approach to Benchmarking Neural Code Generation}

\author{
Federico Cassano,
John Gouwar,
Daniel Nguyen,
Sydney Nguyen,
Luna Phipps-Costin,
Donald Pinckney,
Ming-Ho Yee,
Yangtian Zi,
Carolyn Jane Anderson,
Molly Q Feldman,
Arjun Guha,
Michael Greenberg,
Abhinav Jangda\textsuperscript{\textsection}
}

\begin{document}

\maketitle
\begingroup\renewcommand\thefootnote{\textsection}
\footnotetext{Authors are listed alphabetically with students first, then faculty.}
\endgroup

\begin{abstract}
Large language models have demonstrated the ability to 
generate both natural language and programming language text. Such models open up the possibility of multi-language code generation: could code generation models generalize knowledge from one language to another? Although contemporary code generation models can generate semantically correct Python code, little is known about their abilities with other languages. We propose \system, a system for translating unit test-driven code generation benchmarks to new languages. We create the first massively multilingual code generation benchmark by using \system to translate two popular Python code generation benchmarks to 18 additional programming languages.

We use \system to extend the HumanEval benchmark \citep{chen2021evaluating} and MBPP benchmark \citep{austin2021program} to 18
languages that encompass a range of programming paradigms and
popularity. Using these new parallel benchmarks, we evaluate the multi-language performance of three state-of-the-art code generation models:
Codex \citep{chen2021evaluating}, CodeGen \citep{salesforce-codegen} and InCoder \cite{fried2022incoder}. We find
that Codex matches or even exceeds its performance on Python for several other languages. The range of programming languages represented in \system allow us to
explore the impact of language frequency and language features on model
performance. Finally, the \system approach of compiling code generation
benchmarks to new programming languages is both scalable and extensible, making it straightforward to evaluate new models, benchmarks, and
languages.
\end{abstract}

\section{Introduction}

\emph{Code generation models}, also known as large language models (LLMs) of code, are deep neural networks trained on massive corpora of source code.
Over the past few years, code generation models have demonstrated their utility on a wide variety of software engineering tasks, including test generation, documentation generation, and even synthesizing working programs from natural language descriptions~\citep{gpt-neoX,chen2021evaluating,fried2022incoder,nijkamp2022conversational,xu2022systematic}.
New products such as GitHub Copilot\footnote{\url{https://github.com/features/copilot/}}, Amazon CodeWhisperer\footnote{\url{https://aws.amazon.com/codewhisperer/}}, and Tabnine\footnote{\url{https://www.tabnine.com/}} built on code generation models are growing in popularity with developers~\cite{ziegler2022productivity}.
Although several code generation models are trained on multiple programming languages, they are typically only evaluated on a single programming language: Python. Machine learning researchers are familiar with Python: they have painstakingly constructed several Python code generation benchmarks~\citep{kulal2019spoc,hendrycks2021measuring,austin2021program,chen2021evaluating} and it is the best represented language in training datasets~\cite{yin2018learning,chen2021evaluating,austin2021program,hendrycks2021measuring}.
However, we should also evaluate code generation models with other languages to support a wider variety of programmers.

In this paper we present \system, a system for translating code generation benchmarks from Python into new languages, and use it to propose the first massively parallel, multi-language benchmark for code generation.
By ``multi-language'' we mean multiple programming languages: \system{} supports \nlang{} languages and is straightforward to extend with more.
By ``parallel'', we mean that \system{} produces parallel problems for each language, thus we can measure performance of a code generation model on a consistent set of problems across multiple programming languages.
What makes \system{} possible is that code generation benchmarks have unit tests to determine if the generated function behaves correctly. However, existing benchmarks only evaluate performance on a single language.

\system{} uses a suite of \nlang{} little compilers from Python benchmarks to each target language.
However, what makes this scale is that these are \emph{not} full-fledged compilers.
Each compiler must be able to translate four components from Python: (1)~a function signature (name and arguments),  (2)~simple unit tests, (3)~a comment describing the expected function behavior, and (4)~type annotations if the target language is statically typed.
Notably, the compiler does not have to translate the body of a function, since it is the job of the code generation model to synthesize it.
Thus each \system{} compiler is approximately 200 LOC and easy to build.
\system{} also includes a simple, rule-based tool to translate  technical terms in comments to be more language appropriate, e.g. a Python list is approximately a C++ vector.

\system also includes a containerized sandbox that (1)~compiles programs if necessary, (2)~runs them with appropriate timeouts, (3)~validates their results on unit tests, and (4)~classifies each output as successful, syntax error, etc. Thus each language requires an evaluation script, which is typically about 20 LOC.

We use \system{} to translate two widely-used code generation benchmarks, HumanEval \citet{chen2021evaluating} and MBPP \citet{austin2021program}, into \nlang{} languages. The \nlang{} languages capture a broad spectrum of language features, application areas, and popularity, allowing us to explore the impact of these factors on model performance. 

We use the multi-language parallel \newHE and \newMBPP benchmarks to evaluate three state-of-the-art code generation models:
Codex~\citep{chen2021evaluating}, CodeGen~\citep{salesforce-codegen}, and  InCoder~\citep{fried2022incoder}. Our evaluation presents new insights into the effectiveness of code generation models, including:

\begin{enumerate}
    
    \item Across models and benchmarks, code generation models perform extremely well on JavaScript, sometimes outperforming Python, even on benchmarks originally designed to evaluate Python performance. Codex also performs well on C++, Scala, and TypeScript.

    \item There is no strong correlation between model perplexity and correctness of generated code, which suggests that perplexity may not be a good estimate of performance.

    \item Code generation performance is correlated with language popularity, but some niche languages perform as well as more popular languages.

    \item Code generation performance is sensitive to prompt design for both niche and popular languages.
    
    \item Static type-checking neither helps nor hinders code generation model performance.
    
\end{enumerate}

To summarize, our key contributions are:

\begin{itemize}
    \item \system{}: a suite of compilers and an evaluation framework for translating code generation benchmarks from Python into other programming languages. \system{} translates unit tests, doctests, Python-specific terminology, and type annotations. 
    
     \item Two parallel benchmarks for code generation in 19 languages encompassing a variety of programming paradigms, language features, and popularity levels.
     
    \item A multi-language parallel evaluation of three models, Codex \citep{chen2021evaluating}, InCoder \cite{fried2022incoder}, and CodeGen \citep{salesforce-codegen}.
    
    \item Explorations of language frequency effects, the impact of type annotations, and prompt translation sensitivity on code generation performance, along with a fine-grained error analysis for four languages. 

\end{itemize}

We hope this evaluation work will help the many software engineers that do not use Python to assess the feasibility of code generation models for their work and to understand the language factors that may affect model performance.

\ifanon
\else

\ifsubmission
\system{} is presently being employed to evaluate code generation models for the BigCode project, which is an open collaborative effort to build code generation models on permissively licensed code.\footnote{\url{bigcode-project.org}} 
\fi

The \system{} system, dataset, and tutorial are available at \url{github.com/nuprl/MultiPL-E}.

\fi

\section{Code Generation}\label{sec:task}

Code generation has long been a task of interest: there is extensive work on program synthesis~\cite{alur:sygus,chaudhuri:neurosymbolic, gulwani2017program} using both symbolic and neuro-symbolic approaches. More recently, large neural language models (LLMs) trained for text generation have demonstrated the ability to perform program completion~\citep{brown2020language,gptj,gpt-neoX}.
Since 2020, several code generation models have been trained or fine-tuned on programming language text~\citep{feng2020codebert,clement-etal-2020-pymt5,chen2021evaluating,nijkamp2022conversational,fried2022incoder,xu2022systematic}. 

Unlike traditional program synthesis techniques, neural language models are able to condition on and generate both natural language (i.e., code comments) and programming language text. Moreover, code generation models offer the promise of synthesizing knowledge gleaned from code in multiple programming languages. \textit{Just as language models exposed to multiple natural languages are able to generalize across languages, might not multi-language models of code do the same?}

Although this kind of multi-language generalization is an intriguing possibility, little is known about how well code generation models perform across programming languages. We make progress towards answering this question by proposing two large-scale parallel benchmarks for code generation in 19 languages, which we use to evaluate three state-of-the-art models: Codex, CodeGen, and InCoder.

\subsection{The Natural Language to Code Task}

\begin{figure}[t]
\begin{subfigure}{\columnwidth}
\caption{The prompt given to the NL2Code model:}
\label{prompt-example}
\begin{lstlisting}[
  language=Python,
  linebackgroundcolor={%
  \ifnum\value{lstnumber}=1\color{signaturecolor}\fi
  \ifnum\value{lstnumber}=10\color{doctestcolor}\fi  
  \ifnum\value{lstnumber}=11\color{doctestcolor}\fi
  \ifnum\value{lstnumber}=12\color{doctestcolor}\fi
}]
def lsi(lst):
  """Create a function that returns a 
  tuple (a, b), where 'a' is the largest
  of negative integers, and 'b' is the
  smallest of positive integers in a 
  list. If there is no negative or 
  positive integers, return them as 
  None.
  Examples:
    lsi([2, 4, 1, 3, 5, 7]) == (None, 1)
    lsi([]) == (None, None)
    lsi([0]) == (None, None)
  """
\end{lstlisting}
\end{subfigure}

\tikz\draw [dash dot] (0,0) -- (8,0);

\begin{subfigure}{\columnwidth}
\vspace{1em}
\caption{A completion generated by Codex:}
\label{generated-code-example}
\begin{lstlisting}[
  language=Python
]
  if len(lst) == 0:
    return (None, None)
  else:
    neg_nums = [x for x in lst if x<0]
    pos_nums = [x for x in lst if x>0]
  if len(neg_nums) == 0:
    neg_nums = None
  if len(pos_nums) == 0:
    pos_nums = None
  return (max(neg_nums), min(pos_nums))
\end{lstlisting}

\end{subfigure}

\tikz\draw [dash dot] (0,0) -- (8,0);

\begin{subfigure}{\columnwidth}
\vspace{1em}
\caption{3 of the 12 unit tests (the generated solution fails two):}
\label{unit-tests-example}
\begin{lstlisting}[language=Python, mathescape=true]
$\makebox[0.5em]{\textcolor{red}X}$ assert lsi([2, 4, 1, 3, 5, 7]) == (None, 1)
$\makebox[0.5em]{\textcolor{red}X}$ assert lsi([2, 4, 1, 3, 7, 0]) == (None, 1)
$\makebox[0.5em]{\textcolor{green}\checkmark}$ assert lsi([1, 3, 4, 5, 6, -2]) == (-2, 1)
\end{lstlisting}
\end{subfigure}
\caption{Problem 136 of 164 of the HumanEval benchmark. We shorten the name
\texttt{largest\_smallest\_integers} for brevity. Top: the prompt for the model, with the
\bgroup\fboxsep=1.5pt
\colorbox{signaturecolor}{function signature}, natural language description, and \colorbox{doctestcolor}{doctests}.
\egroup
Middle: a Codex-generated solution. Bottom: unit tests.}
\label{fig:prompt_examples}
\end{figure}

\noindent
Code generation models have been applied to a variety of tasks, including test generation \citep{tufano2020unit}, docstring generation \citep{lu2021codexglue}, code search \citep{feng2020codebert,ahmed2022multilingual}, type inference~\cite{wei:lambdanet,hellendoorn:dl-ti,pradel:typewriter}, and more \cite{drori:codex-univ-math}. 
We focus on the \textbf{natural-language-to-code} task (NL2Code): given the description of a function in natural language, complete the function body.

The input to a code generation model is called a \emph{prompt}.
\Cref{prompt-example} shows an example prompt from the HumanEval benchmark for NL2Code~\citep{chen2021evaluating}. The prompt has several sources of information for the model: the function signature (its name and parameters); a brief comment describing the function; and, optionally, examples in the form of Python doctests.
Given the prompt as input, the code generation model generates a \emph{completion} that is likely to follow the given prompt.

Note that the model does not receive an explicit cue about the target language, but each of the three prompt regions provide implicit cues: the syntax of the function signature, the terminology used in the natural language description, and the syntax of the doctests all suggest that the target is Python. Consequently, to translate this prompt to a new programming language, we must target all three regions of the prompt.

\subsection{Sampling Program Completions}\label{subsec:sampling}

There are several ways to configure how a code generation model produces completions, each of which can have a significant effect on the quality of generated code.
Fundamentally, a completion is a sequence of tokens and is \emph{not} an abstract syntax tree.
Therefore, a completion can readily produce tokens that go beyond a single function.
For example, given just the the signature of ``mean'', InCoder produces the mean, variance, standard deviation, and several other functions (\cref{incoder-long-output}).
In fact, it can continue producing code up to the maximum sequence length, which, for InCoder, is 2048 tokens.

We control this output by specifying \emph{stop sequences} that typically demarcate the end of a function.
For Python, we use the stop sequences that have been employed in prior work~\cite{chen2021evaluating}. For example, when completing a top-level function, \lstinline|\ndef| marks the start of the next top-level function, but allows nested helper functions. For other languages, we design different sets of stop sequences (\cref{app:lang-eval}).

Under the hood, given a prompt, a code generation model produces a completion one token at a time.
At each step, the neural network receives an encoded prompt as input and produces a distribution for the following token.
To generate several tokens, a \textit{sampling algorithm} iteratively samples next tokens, extending the prompt at each step with the previously sampled token.

There are a variety of sampling approaches that one can use.
A naive approach is to greedily sample the next most likely token, but this performs poorly in practice~\cite{text-degeneration}. One approach employed in prior work~\citep{chen2021evaluating}, and in this article, is to rescale the probability distribution to favor high probability tokens more strongly using a \textit{temperature} hyperparameter ($0 \le t < 1$): low temperature makes the completion more ``predictable'' and high temperature makes it more ``creative''. This is commonly combined with top-$p$ sampling, which cuts off the least likely tokens that contribute in aggregate $1-p$ to the probability mass, and redistributes their mass to the remaining tokens.

\subsection{Evaluating Code Generation}

Early work on code generation relied on textual similarity metrics for evaluation~\cite{feng2020codebert,ren2020codebleu}. However, previous work shows that textual similarity is not reliably correlated with code correctness~\cite{austin2021program,chen2021evaluating}. The best way to evaluate code generation is to test code correctness using a suite of hidden unit tests.

We translate two code generation benchmarks that include unit tests for every problem. \Cref{unit-tests-example} shows 3 of the 12 unit tests that accompany the problem from \Cref{prompt-example}. Note that these unit tests are simple assertions: each test asserts that the output value produced by the function matches an expected value. 

We judge a generated function correct if it passes \emph{all tests}.
\Cref{generated-code-example} shows just one of the solutions generated by Codex for the example prompt. This solution is incorrect because it fails some of the unit tests (\cref{unit-tests-example}). Because the output of the code generation model is stochastic, it is common to sample multiple completions per problem and report an estimated pass rate (\Cref{subsec:metrics}).

\begin{figure}
\begin{lstlisting}[
  language=Python,
  linebackgroundcolor={%
  \ifnum\value{lstnumber}=1\color{signaturecolor}\fi
}]
def mean(n):
    return sum(n)/len(n)

def variance(n):
    mean = mean(n)
    return sum([(n-mean)**2 for n in n])/len(n)

def standard_deviation(n):
    return math.sqrt(variance(n))

def mode(n):
    counts = Counter(n)
    max_count = max(counts.values())
    return [k for k,v in counts.items() 
            if v == max_count]
\end{lstlisting}
\caption{Code generation models produce tokens, not ASTs, and may produce output beyond that requested. This is truncated output from InCoder given
just the first highlighted line as the prompt.}
\label{incoder-long-output}
\end{figure}

\section{The \system{} Approach}\label{sec:dataset}

This section describes how we select and prepare languages and benchmarks for \system{}.

\subsection{Benchmark Selection}

There are a number of existing single-language NL2Code benchmarks \citep{yin2018learning,kulal2019spoc,hendrycks2021measuring}. 
We choose to translate HumanEval~\citep{chen2021evaluating} and MBPP~\cite{austin2021program} as two of the most widely-used benchmarks.

HumanEval is a good choice of benchmark for several reasons. It is a diverse collection of 164 problems, where all problems have tests to check correctness, and most have examples or doctests as part of the prompt. All of the problems are functions that receive and return first-order values, which facilitates unit testing and test translation. Many also use Python's optional type annotations. Moreover, it is a challenging benchmark: the best model evaluated by Fried et al. \citet{fried2022incoder} achieves only a 36\% pass rate on Python.

MBPP is another large, commonly used benchmark of Python problems. As originally formulated, it is a little unusual.
Each problem has a description and a list of assertions.
The prompt for code generation includes both the description and the assertions, and the generated code is then tested with the same set of assertions.
We argue that the HumanEval approach, where test cases are hidden, is a significantly better way to evaluate code generation. We therefore remove the assertions from the MBPP prompts so that we can use them as hidden unit tests.
However, with only a problem description, a code generation model is free to make up the name of a function (or not even produce a function).
Therefore, we mechanically augment every prompt with a function signature, based on the name and arity implied by the assertions. \Cref{mbpp-modifications} shows an example of an original MBPP prompt and our modification.

\begin{figure}
\begin{subfigure}{\columnwidth}
\begin{lstlisting}[language=Python]
# Write a function to fnd the smallest missing
# element in a sorted array. Your code should
# satisfy these tests:
assert smallest_missing([0, 1, 2, 3, 4, 5, 6], 
                        0, 6) == 7
assert smallest_missing([0, 1, 2, 6, 9, 11, 15],
                        0, 6) == 3
assert smallest_missing([1, 2, 3, 4, 6, 9, 11, 15],
                        0, 7) == 0
\end{lstlisting}
\caption{An original MBPP prompt: the same assertions are used to test the generated code. (Typo in comment is from the original benchmark.)}
\label{original-mbpp}
\end{subfigure}

\vspace{2em}

\begin{subfigure}{\columnwidth}
\begin{lstlisting}[language=Python]
def smallest_missing(l):
   """
   Write a function to fnd the smallest missing 
   element in a sorted array.
   """
\end{lstlisting}
\caption{We add the function signature and hide the test cases to do a more rigorous evaluation.}
\label{modified-mbpp}
\end{subfigure}
\caption{An original MBPP prompt and how we modify it to standardize evaluation.}
\label{mbpp-modifications}
\end{figure}

\subsection{Programming Language Selection}

\system{} supports 19 programming languages, which we categorize into four frequency classes (\niche, \low, \med, or \high) based on a weighting of TIOBE rank and GitHub frequency (Table~\ref{tab:pls}).
Eight of the languages in \system{} had never been used before to measure NL2Code performance; this set includes newer languages (Julia and Swift), older scripting languages (Bash and Perl), and languages for specific applications (Lua and R). Half of the languages are statically type-checked. The broad range of languages in \system shows the generality of our compilation approach and allows us to explore how language frequency and language features affect performance (\cref{sec:factors}).

A key feature of \system{} is that it is easy to extend with new models, benchmarks, and languages. To support new languages and benchmarks without manual (and error-prone) effort, we build \nlang{} compilers to translate NL2Code benchmarks written in Python.
Writing one of these compilers is straightforward when the target language is similar to Python, but requires care for typed languages and even some untyped languages, notably Perl, Bash, and R.

\begin{table}[t]
\centering
\begin{tabular}{|l|c|l|l|l|r|}\hline
     PL &Typed? & GitHub \% & TIOBE & Category & LOC \\\hline
          Bash&$\times$&-&43&\niche&120\\
     C++&\checkmark&7.0&4&\high&244\\
      C\#&\checkmark&3.1&5&\med&149\\
       D&\checkmark&-&35&\niche&117\\
     Go&\checkmark&7.9&12&\med&210\\
     Java&\checkmark&13.1&3&\high&153\\
     JavaScript&$\times$&14.3&7&\high&45\\
      Julia&$\times$&0.1&28&\niche&125\\
      Lua&$\times$&0.2&25&\niche&43\\
     Perl&$\times$&0.3&17&\low&49\\
    PHP&$\times$&5.3&11&\med&50\\
     R&$\times$&0.05&19&\low&98\\
     Racket&$\times$&-&-&\niche&38\\
   Ruby&$\times$&6.2&15&\med&41\\
    Rust&\checkmark&1.1&22&\low&147\\
    Scala&\checkmark&1.7&32&\low&152\\
    Swift&\checkmark&0.7&10&\low&479\\
    TypeScript&\checkmark&9.1&33&\high&117\\\hline
\end{tabular}
\caption{\system languages by frequency, as calculated by GitHut 2.0 and the 2022 TIOBE Programming Community index; the LOC column indicates the number of semantic lines of code in our compiler.} \label{tab:pls}
\end{table}

\subsection{Compiling Python Benchmarks}
\label{compilers-lol}


A \system{} compiler is significantly easier to build than a complete compiler. To translate a benchmark problem, we only need to compile function signatures and unit tests
(not arbitrary statements and expressions).
Our compilers preserve comments, since they contain the natural language description for the NL2Code task; however, we automatically rephrase them to replace Python-specific terminology.

\subsubsection{Compiling Unit Tests}
\system{} supports any unit test where the input and output to the test are \emph{first-order values}. In Python, these include constants and data structures such as lists, tuples, and dictionaries, but exclude values such as \texttt{lambda} expressions.\footnote{We do not support testing higher-order functions, but support generated code that uses higher-order functions.} HumanEval and MBPP unit tests apply the model-generated function to a first-order value, and compare the result with an expected first-order value.

Each \system{} compiler has a recursive function that compiles Python values to the target language's values.
Even for an untyped target, this value-to-value compilation requires care, because not all Python value types have perfect analogues in every target.
For example, we compile both tuples and lists to JavaScript arrays, since JavaScript lacks a canonical tuple type.
We also support untyped targets where the compilation strategy is less obvious. For example, when the target is R, it may appear natural to compile Python lists to R lists: both kinds of lists can be nested and allow heterogenous values.
However, R's vector type is much more commonly used (data frames are made of vectors).
Unfortunately, vectors must be homogeneous and cannot be nested, so not all Python lists can be translated to vectors.
For example, an argument typed \lstinline|List[Int]| can be translated to a vector, but a nested list cannot. In order to more closely match the token distribution of idiomatic R code seen during training, our R compiler uses types (described below) to identify homogenous list values and maps them to vectors using \lstinline|c()|---even though R is untyped.

The final step of compiling tests is to choose an appropriate test for equality.
The meaning of equality operators varies across programming languages.
Python's \lstinline|==| operator checks \emph{deep equality}, i.e., item-by-item equality within data structures. Deep equality is the appropriate choice for unit tests. In some languages, we need to import equality-testing functions
from testing libraries, as in the JavaScript example shown in \Cref{fig:unit-tests}.

\begin{figure}
\begin{subfigure}{\columnwidth}
\caption{Original Python assertion.}
\begin{lstlisting}[language=python]
assert lsi([0]) == (None, None)
\end{lstlisting}
\end{subfigure}

\vspace{0.5em}
\tikz\draw [dash dot] (0,0) -- (8,0);
\vspace{0.5em}

\begin{subfigure}{\columnwidth}
\caption{Equivalent R.}
\begin{lstlisting}{r}
if(!identical(lsi(c(0)), c(NULL, NULL))){
    quit('no', 1)}
\end{lstlisting}
\end{subfigure}

\vspace{0.5em}
\tikz\draw [dash dot] (0,0) -- (8,0);
\vspace{0.5em}

\begin{subfigure}{\columnwidth}
\caption{Equivalent JavaScript.}
  
\begin{lstlisting}
assert.deepEqual(lsi([0]), [void 0, void 0]);
\end{lstlisting}
\end{subfigure}
    
\caption{Example of a translated assertion.}
\label{fig:unit-tests}
\end{figure}

\subsubsection{Translating Types and Type Inference}
%
%
Compiling a function signature to an untyped language is straightforward, but requires care when the target is typed. Most typed languages require argument and return type annotations.
Python has optional type annotations.
Thus if a benchmark has type annotations, we can translate them to types in a target language.
Fortunately, a large subset of the HumanEval benchmarks employ Python's optional type annotations. We introduce type annotations to the few that do not.
None of the MBPP benchmarks have type annotations.
Instead of manually adding annotations to 400+ benchmarks, we infer the types of the values that appear in the MBPP assertions.

Translating types and typed values is subtly different for every language.
For example, five HumanEval problems use types such as \texttt{Any} which cannot be translated to most traditional statically typed languages (e.g., C++ and Rust). We fail to compile these few problems to these languages.

Another problem arises when compiling to languages with algebraic datatypes or discriminated unions.
For example, consider translating the Python type \lstinline|Optional[Int]| to Rust, Swift, or Scala.
The analogous type in the target language is an algebraic datatype.
This means that when the Python number \lstinline|n| has type \lstinline|Optional[Int]| it must translate to the value \lstinline|Some(n)|.
Optional values are very common in Python benchmarks, and we use this approach extensively.

Finally, many typed languages require type annotations in data structures, which appear in unit tests. For example, C++ vectors require an annotation specifying their element type, and numbers  in Rust (sometimes) require a type suffix. We perform limited local type inference to calculate these types from the type of the function signature to ensure that the unit tests always compile successfully.

\subsubsection{Translating Doctests}
Python \emph{doctests} are a standard format for examples in documentation.
While many of the HumanEval prompts include examples, not all of them are validly formatted doctests. We standardize examples to the Python doctest format ("\textgreater{}\textgreater{}\textgreater" prepended). We apply value-to-value compilation to the doctests as we do for unit tests. However, since not all languages have an equivalent doctest format, we keep the Python format for all target languages.

\subsubsection{Translating Python Terminology in Prompts}

Different programming languages use different terminology to refer to the same concept. For example,
a Python \emph{list} is closest to a JavaScript \emph{array} or a Rust \emph{vector}.
To mitigate the impact of these differences, we identify Python-specific terminology in the natural language portion of the prompt, and translate it to the most natural equivalent for the target language.
Figure~\ref{fig:lang-in-prompts} shows an example of a prompt translated from Python to Perl. Notably, Perl not only lacks Booleans, but uses 1 for true and the empty string for false.

\begin{figure}[t]
\begin{subfigure}{\columnwidth}
\caption{Original Python docstring from HumanEval \#95.}
\footnotesize
\texttt{
Given a \colorbox{red!40}{dictionary}, return \colorbox{pink!20}{True} if all keys are strings in lower case or all keys are strings in upper case, else return \colorbox{pink!20}{False}. The function should return \colorbox{pink!20}{False} is the given \colorbox{red!40}{dictionary} is empty.}
\end{subfigure}
\vspace{0.5em}
\begin{subfigure}{\columnwidth}
\caption{Terminology translated to Perl.}
\footnotesize
\texttt{
Given a \colorbox{red!40}{hash}, return \colorbox{pink!20}{1} if all keys are strings in lower case or all keys are strings in upper case, else return \colorbox{pink!20}{""}. The function should return \colorbox{pink!20}{""} is the given \colorbox{red!40}{hash} is empty.}
\end{subfigure}

\caption{A Python docstring and its Perl translation. Errors (e.g., ``is'' for ``if'') are from the original benchmark. }\label{fig:lang-in-prompts}
\end{figure}

We conservatively avoid translating number types. Although some target languages use different terms for floats and integers, the term \textit{integer} is commonly used in a mathematical sense rather than in reference to the Python type.

\subsection{Limitations of Our Approach}
\label{limitations}

 \begin{figure*}[t]
    \centering
    \includegraphics[width=\textwidth]{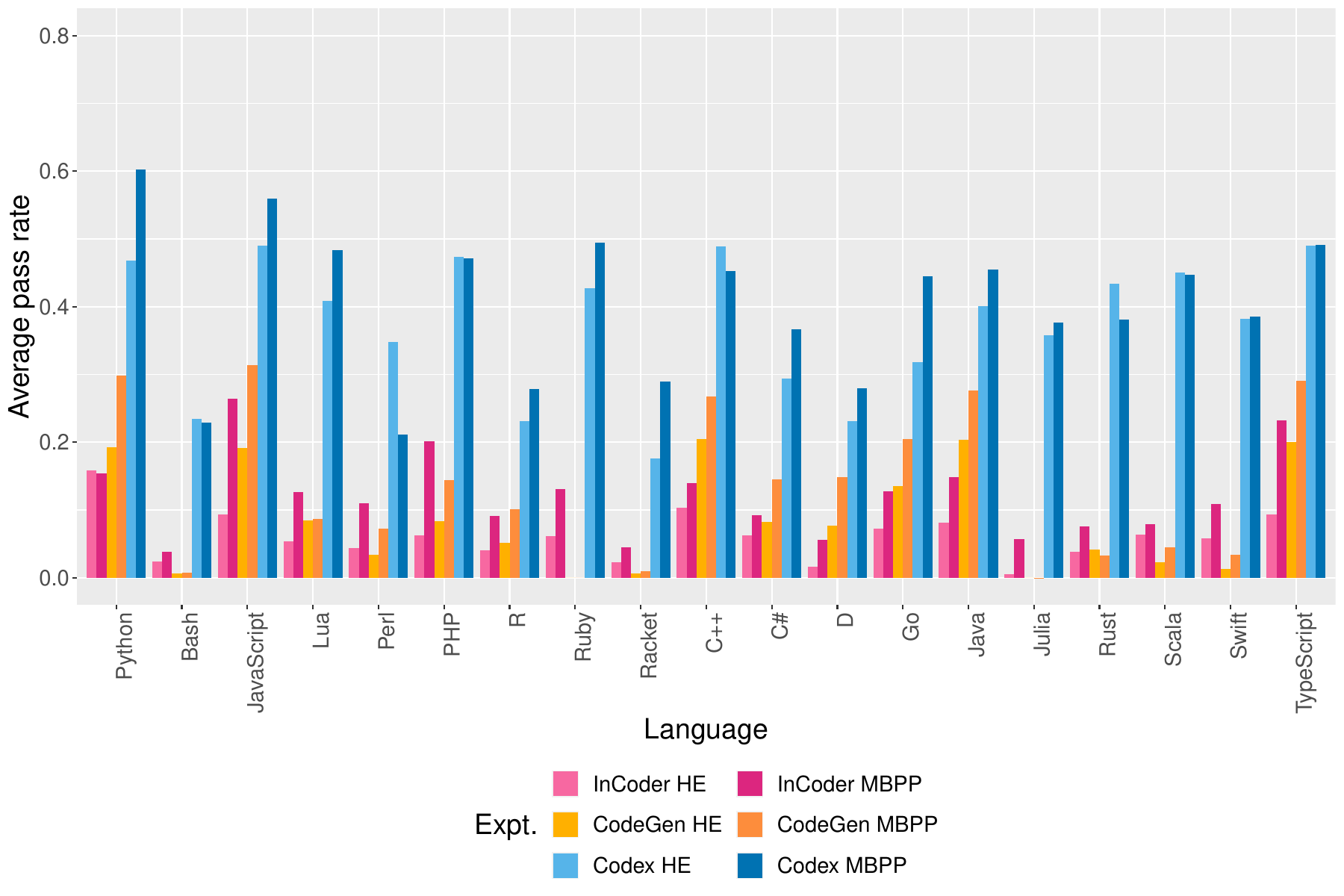}
    \caption{Pass@1 rates for all languages in \newHE and \newMBPP. From left to right: InCoder, CodeGen, Codex.}
    \label{fig:total_results}
\end{figure*}

A handful of benchmarks cannot be easily translated using the \system{} approach. Of the 164 original HumanEval benchmarks:
(1)~we exclude 3 benchmarks that have Python helper functions in their prompt;
(2)~we modify 2 benchmarks to use unit tests instead of randomized testing; and (3) for certain typed languages, we fail to compile up to 5 benchmarks with untranslatable types. These changes do not lead to significantly different results for Python (\Cref{subsubsec:he_replication}).

Our approach can be generalized to additional programming languages, so long as the target language has natural analogues for the Python data types used in the benchmarks. We do not include two previously studied languages, C \cite{xu2022systematic} and SQL \cite{yu-etal-2018-spider,wikisql} because they do not meet this criterion.

\section{Code Generation Models}

We evaluate three state-of-the-art code generation models, each of which use a Transformer architecture~\citep{vaswani2017attention} and are trained with a language modeling objective on a mixture of natural language and code. We evaluate the largest, best-performing versions of each of these three models.  

\subsection{Models}\label{subsec:models}

\newpara{InCoder} InCoder \citep{fried2022incoder} is a 6.7B parameter language model trained using a causal masking objective~\citep{aghajanyan2022cm3}. It supports both code infilling and code completion; we test only the latter. InCoder was trained on 159 GB of deduplicated, filtered code from Github (around a third in Python) and 57 GB from StackOverflow.

\newpara{CodeGen} CodeGen is a 16.1B parameter language model trained with a next-token prediction objective. We evaluate the multilingual CodeGen model, which was trained first on The Pile~\citep{thepile}, a 825 GB dataset of mostly natural language text with around 8\% Github-scraped code. The model was further trained (fine-tuned) on a 6 programming language subset (C, C++, Go, Java, JavaScript,
and Python) of the BigQuery code dataset.\footnote{https://cloud.google.com/blog/topics/public-datasets/github-on-bigquery-analyze-all-the-open-source-code}

\newpara{Codex} Codex is a GPT-3 language model fine-tuned on code. \citet{chen2021evaluating} describe a 12B parameter version of Codex fine-tuned on 159 GB of deduplicated, filtered Python code from Github. We use the more recent \texttt{codex-davinci-002} model, which is trained on multiple languages. Details of its training set and size are not public~\citep{codexblog}. We use the public OpenAI API to query Codex.

\subsection{Metrics}\label{subsec:metrics}

For each language, we calculate pass@$k$ using the methodology employed by 
 \citet{chen2021evaluating} and subsequent work.
 Intuitively, pass@1 is the likelihood of the model producing a completion that passes \emph{all unit tests}, pass@10 is the likelihood of any one of 10 completions passing all unit tests, and so on.
 We calculate pass@1 with temperature $0.2$, and use temperature $0.8$ for pass@10 and pass@100. For statistical reliability, we take 200 completions at each temperature and calculate average pass rate using the unbiased sampling estimator presented in \citet{chen2021evaluating}.\footnote{We note that pass@1 rates appear to stabilize around 20 samples, suggesting that future work could achieve a stable estimate with a less computationally costly sample size.}
 
 \section{Evaluation}\label{sec:results}
 
 In this section, we present the results of evaluating Codex, InCoder, and CodeGen on \newHE and \newMBPP. We fit mixed-effects models to evaluate the statistical significance of the differences between groups that we report below~\citep{lme4}.
 Appendix~\ref{app:mem} has a full description of each statistical model with its estimate table.
 
 \subsection{\newHE results}\label{subsec:newhe-results}
 
 \begin{figure*}[t]
\centering
\includegraphics[width=\textwidth]{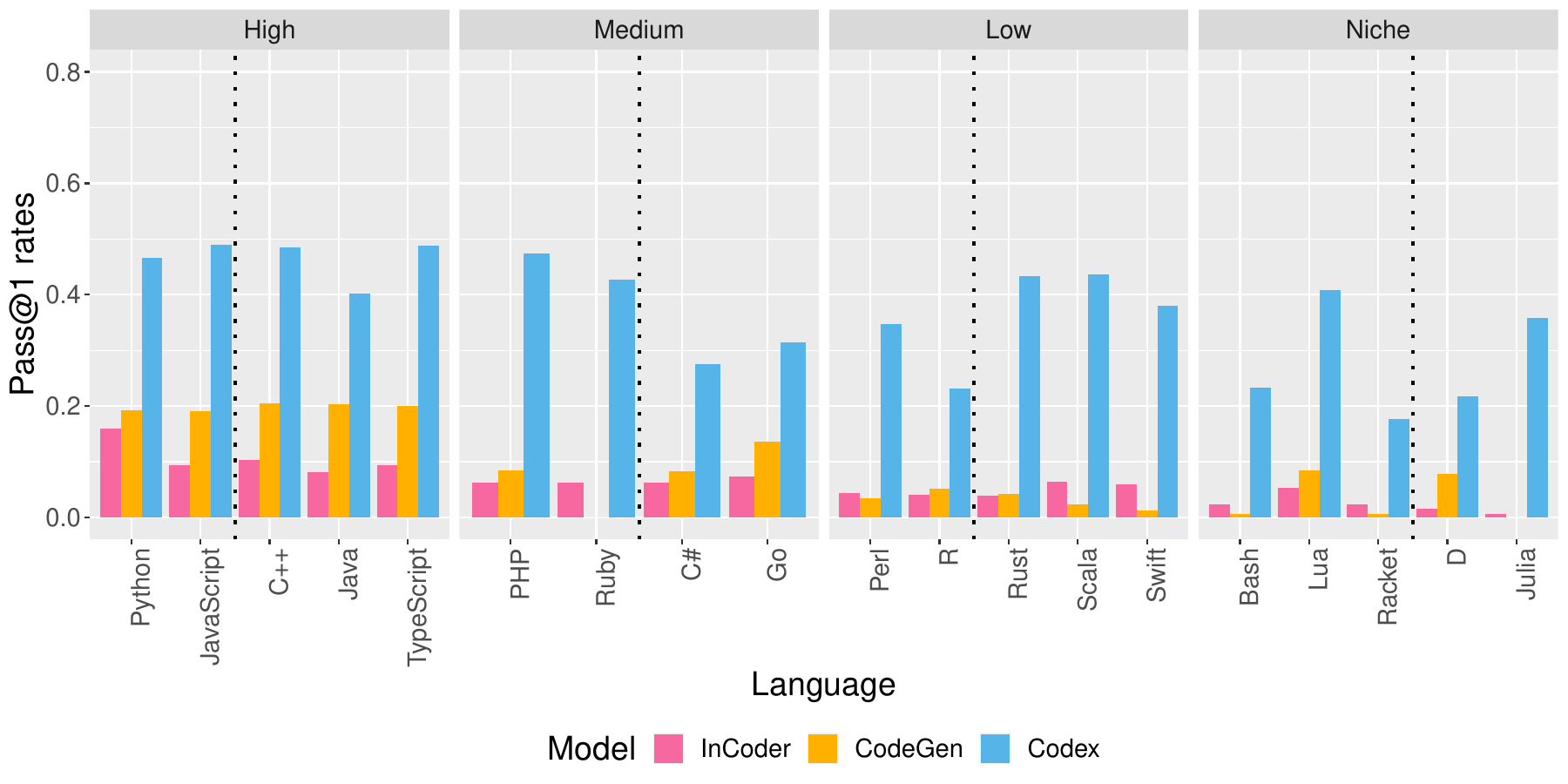}
\caption{Model performance on \newHE by language frequency and type-checking. Languages to the left of dashed line are untyped; languages to the right are  typed.}\label{fig:he-lang-frequency}
\end{figure*}
 
We explore the code generation abilities of the three models on our translated version of HumanEval, \newHE.  Figure \ref{fig:total_results} shows the by-language performance of each model on both benchmarks.

We find reliable differences between Codex pass@1 rates on \newHE for Python and all but 4 languages: C++, JavaScript, Scala, and TypeScript. For these languages, Codex performance is similar to Python.

CodeGen performs best on the languages included in its fine-tuning dataset (Python, JavaScript, Java, C++, and Go). It also performs well on TypeScript, likely due to its similarity to JavaScript. A mixed-effects model finds reliable differences in pass@1 rates on \newHE between all languages and Python, except Ruby, where performance is so low that the model fails to find a reliable estimate.

InCoder performs significantly better on the Python version of \newHE than all of the other language versions ($p<0.001$ for all languages). 

\subsubsection{Python Results and Replication}
\label{subsubsec:he_replication}
Our InCoder results on Python exactly replicate its previously reported performance on HumanEval~\citep{fried2022incoder}. We measure a slightly higher pass@1 rate for CodeGen than what is reported in \cite{salesforce-codegen} (19.2\% compared to 18.3\%).\footnote{We note that \cite{salesforce-codegen} calculates the pass@1 rate for 3 temperatures, and reports the best result without specifying the temperature. Consequently, it's not clear whether the 18.3\% pass@1 rate they report is measured at the 0.2 temperature that we use.}  These findings show that the small standardization changes we made to the HumanEval benchmarks do not significantly affect model performance.

We evaluate a more recent Codex model (\texttt{code- davinci-002}) than the original paper and observe a large improvement on Python: a pass@1 rate of 45.9\%, compared to 28.8\% reported earlier~\cite{chen2021evaluating}. Our pass@1 rate on the Python HumanEval subset replicates what is reported for \texttt{code-davinci-002} in \citet{codeT}.\footnote{We note that \citet{codeT} also reports results for InCoder and a monolingual version of CodeGen, but with sampling differences that the pass@1 rates difficult to compare.}

\subsubsection{Codex Performs Best on JavaScript}
Codex's performance on JavaScript is better than its performance on Python, though the difference is not significant (+2.3\%; $p=0.43$). Codex achieves a pass@1 rate higher than 40\% on C++, Java, TypeScript, PHP, Ruby, Rust, Scala, and Lua. 

The Codex training set is not public; it is possible that the latest model has been trained on solutions to the HumanEval benchmarks in Python, and this could be inflating its performance.
However, \newHE is a new dataset for the \nlang{} other languages. That Codex matches or exceeds its Python performance on some of these new languages suggests a negligible impact of any train/test overlap.

CodeGen also performs well on JavaScript and TypeScript, though the latter is not included in its fine-tuning dataset.

InCoder's performance is the weakest. Like the other models, it performs better on more frequently-used languages (Python) than less popular ones. However, unlike Codex and CodeGen, it does not match its Python performance on any other language.

\subsubsection{Performance by Language Frequency}\label{subsubsec:he_freq}

\Cref{fig:he-lang-frequency} shows \newHE pass@1 rates for each model, grouping languages by frequency. All three models perform best on high frequency languages. 

Although we find reliable differences in Codex pass@1 rates between the \textsc{Medium},  \textsc{Low}, and \textsc{Niche} languages when compared to the \textsc{High} category ($p=0.006$; $p<0.001$; $p=0.002$), we observe that Codex performs very well on some \textsc{Low} and \textsc{Niche} languages. For instance, Lua is the 9th-best language in our Codex evaluation, although it only appears in 0.2\% of GitHub activity and is not in the TIOBE Top-20. CodeGen also performs well on Scala, Rust, and Julia.

Our evaluation therefore shows that contemporary code generation models may be useful even for developers working with less commonly used programming languages.

\subsubsection{Perplexity and Code Correctness Do Not Correlate}

\begin{figure}[t]
    \centering
    \includegraphics[width=0.45\textwidth]{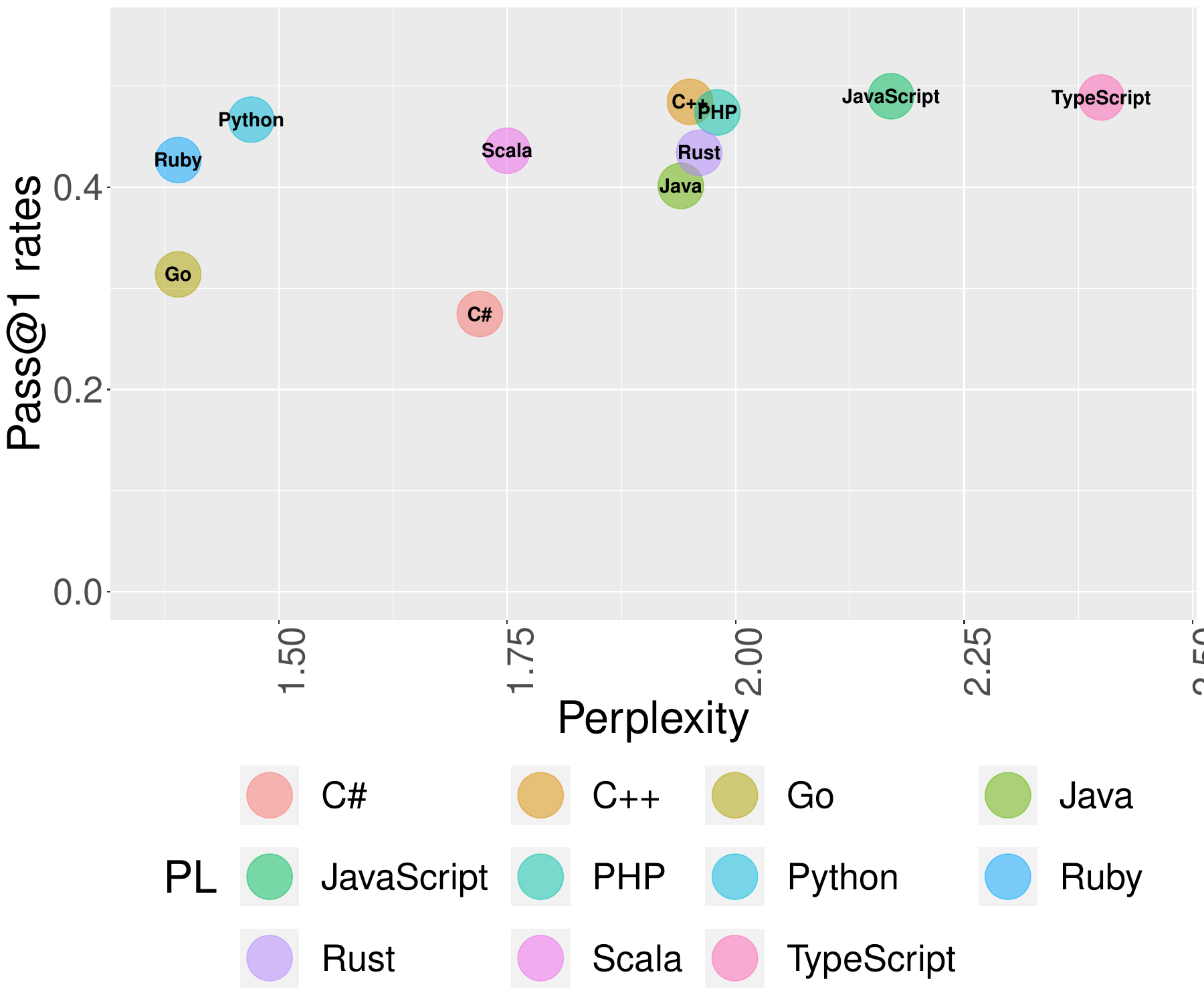}
    \caption{Codex HumanEval pass@1 rates versus perplexity scores reported in \citet{xu2022systematic}.}
    \label{fig:perplexity}
\end{figure}

Xu et al. \citet{xu2022systematic} report Codex perplexity scores for 11 of our 18 languages. We do not observe a strong correlation between Codex pass@1 scores and their perplexity scores (\Cref{fig:perplexity}).
Notably, perplexity is highest for JavaScript and TypeScript, while we find that Codex performs \emph{best} on these languages.
Therefore, perplexity may not be a reliable evaluation metric for NL2Code. One caveat is that \citet{xu2022systematic} likely evaluate an older Codex model, since they report substantially lower pass rates for Python.

\begin{figure*}[t]
\centering
\includegraphics[width=\textwidth]{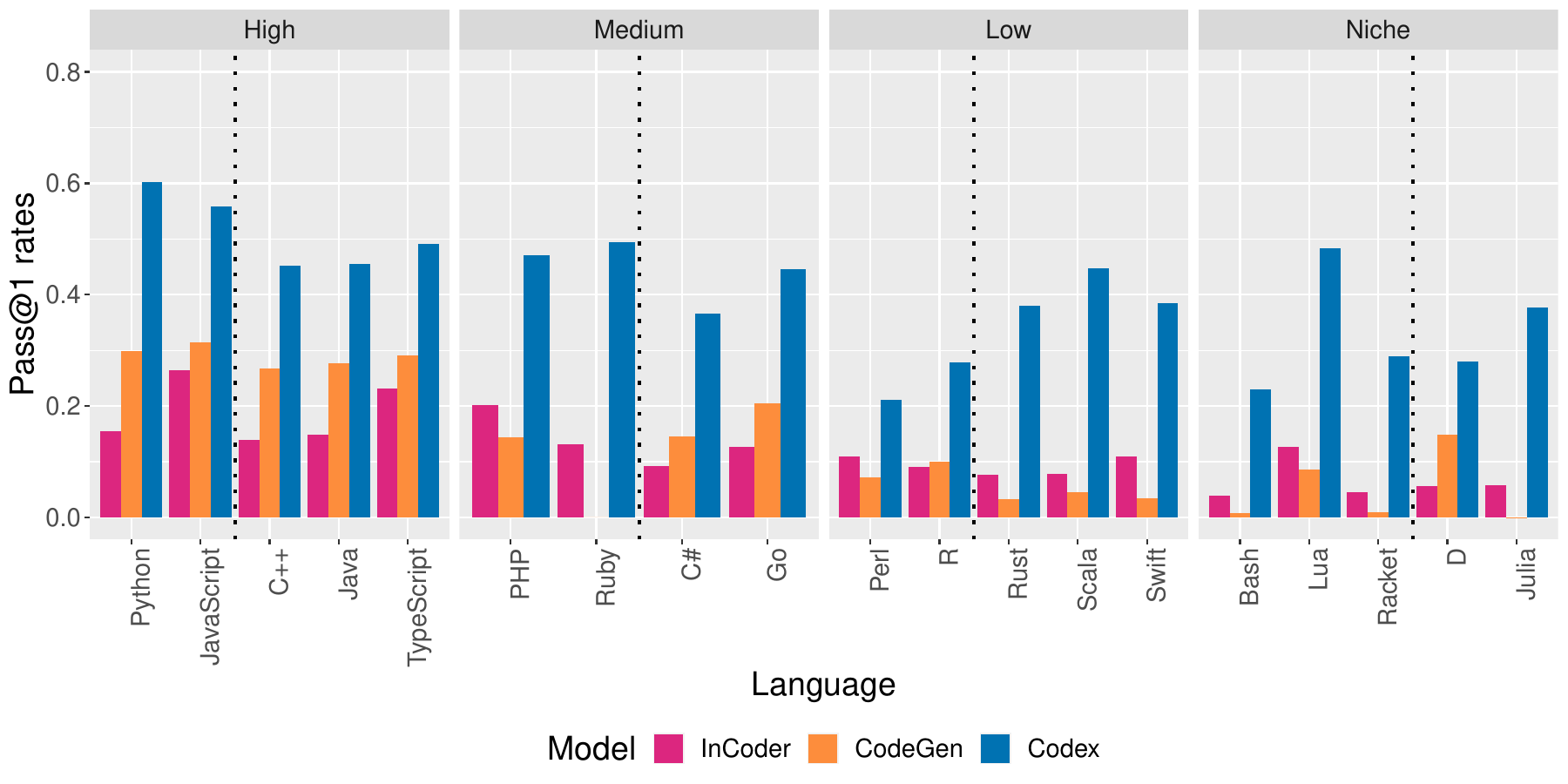}
\caption{Model performance on \newMBPP by language frequency and type-checking. Languages to the left of dashed line are untyped; languages to the right are typed.}\label{fig:mbpp-lang-frequency}
\end{figure*}

\subsection{\newMBPP results}\label{subsec:newmbpp-results}

Figure \ref{fig:total_results} shows the by-language performance for Codex, CodeGen, and InCoder on our translated version of the MBPP benchmark, \newMBPP. Codex performs strongest on the Python problems, but, as with \newHE, does well on several other languages, including JavaScript. A mixed-effects model finds significant differences in Codex pass@1 rates between Python and all other languages in \newMBPP.

As with \newHE, CodeGen performs best on the \newMBPP languages included in its fine-tuning set: Python, JavaScript, C++, Java, and Go. It performs almost as well on TypeScript as on JavaScript. A mixed-effects model finds significant differences in CodeGen pass@1 rates between Python and all languages except Ruby, where performance is so low that the model fails to find a reliable estimate. 

Unlike with \newHE, on \newMBPP, InCoder's performance on TypeScript, JavaScript, and PHP actually exceeds its performance on Python. InCoder's Python pass@1 rate is similar on \newHE and \newMBPP, one of the few instances where MBPP performance is not considerably better than HumanEval. A mixed-effects model finds significant differences in InCoder pass@1 rates for all languages, with positive coefficients for TypeScript, JavaScript, and PHP.

We note that MBPP, unlike HumanEval, does not include any doctests in the prompts. The observed differences in performance on \newMBPP and \newHE in certain languages may relate to this, as we discuss in more detail in \Cref{subsec:ablation}.  

\subsubsection{MBPP is Less Challenging Than HumanEval}\label{subsubsec:less_challenging}

MBPP appears to be a less challenging benchmark than HumanEval. The \newMBPP pass@1 rate is higher than the \newHE pass@1 rate for all but 6 of our 57 model/language pairs. This is despite the fact that MBPP does not provide doctests in any prompts, which, as we show in \Cref{subsec:ablation}, affects performance for some languages.

This suggests that HumanEval may be a more useful benchmark suite than MBPP for the community, as it provides an equally good or better indication of model performance with a more computationally efficient sample size.

\subsubsection{Python Results and Replication}\label{subsubsec:mbpp_replication}

Our Python MBPP pass@1 rates for Codex are slightly higher what is reported in \cite{codeT} (60.3\% compare to 58.1\%). \citet{codeT} prompts with a function signature and docstring, even though the original MBPP problems do not include function signatures; we also include function signatures, which we infer from the provided test cases.

Our Python MBPP results for InCoder are lower than what is reported in the original paper (15.5\% compared to 19.4\%)~\citep{fried2022incoder}. We calculated pass@1 rates for MBPP differently than Fried et al. \citet{fried2022incoder} in two key ways. First, since the original MBPP benchmarks do not include function signatures, Fried et al. \citet{fried2022incoder} prompts InCoder with the MBPP docstring only. We infer function signatures for MBPP problems from the provided test cases, as described in \Cref{sec:dataset}. Second, Fried et al. \cite{fried2022incoder} reports computing pass@1 rates for MBPP using a single completion, rather than computing the unbiased sampling estimator with 200 samples as described in Chen et al. \citet{chen2021evaluating}, as we do. We suspect this leads to inflated pass@1 rates.\footnote{With a single sample, the pass@1 rate for an individual problem will be 0\% or 100\%. Assuming that the 200 samples are fairly homogeneous but not identical, as we observe is usually the case, the unbiased sampling pass@1 rate for an individual problem is rarely 100\%, since this would require all 200 samples to be correct.} 

\subsubsection{Performance by Language Frequency} \label{subsubsec:mbpp_freq}

Figure \ref{fig:mbpp-lang-frequency} shows model performance on \newMBPP by language frequency. As with the \newHE benchmark, models generally perform better on more common languages. However, Codex performance on \newMBPP is robust on many \textsc{Medium}, \textsc{Low}, and \textsc{Niche} languages, such as Lua and Scala. CodeGen performs surprisingly well on the D version of MBPP, a niche language not included in its fine-tuning dataset.

We find reliable differences in Codex pass@1 rates between languages in the \textsc{Medium}, \textsc{Low}, and \textsc{Niche} categories when compared to the \textsc{High} category ($p=0.007$; $p<0.001$; $p<0.001$).

\subsection{Summary}

On the whole, our results replicate previously reported model performance on code generation for Python. We benchmark three state-of-the-art models on 18 additional languages, most of which have never been evaluated before. Surprisingly, we find remarkably good model performance on some lower-frequency languages, such as Lua. We also find that performance on JavaScript and TypeScript is consistently high and occasionally exceeds Python, even though the benchmarks we explore originated in Python.

\section{Factors in Code Generation Success}\label{sec:factors}

In this section, we explore factors that impact code generation success. Focusing specifically on the \newHE benchmark suite, we report results from a number of follow-up experiments, including an ablation study of \system's translation components and finer-grained examinations of language features and prompt translation choices. We also provide a fine-grained analysis of the kinds of errors that arise in NL2Code across several languages.

\subsection{Ablation Study}\label{subsec:ablation}

 \begin{figure*}[t]
\centering
\includegraphics[width=\textwidth]{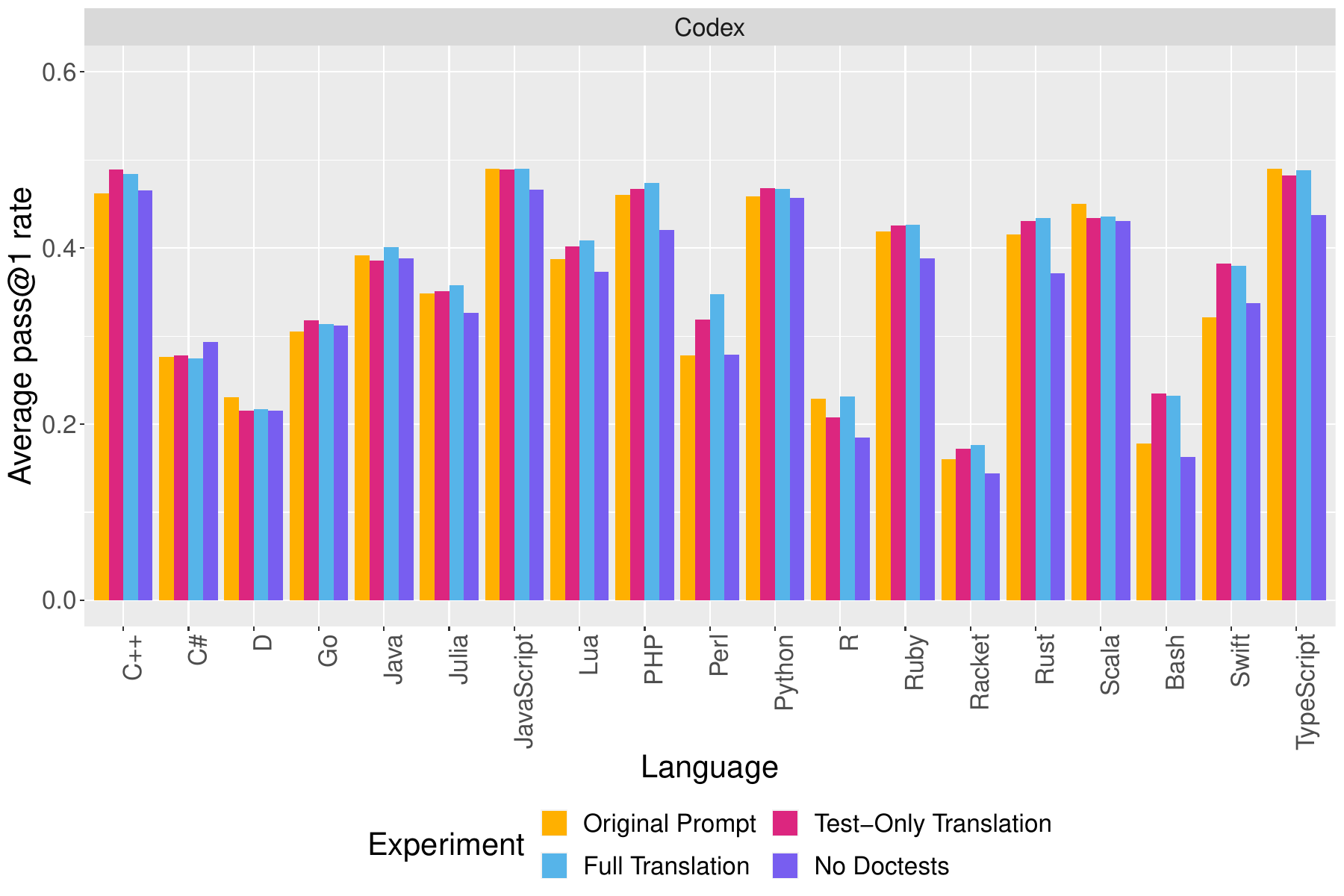}
\caption{Ablation study of translation components, showing Codex pass@1 with original prompts; translated doctests; translated text and doctests; and doctests removed.}\label{fig:ablation}
\end{figure*}

Our compilers target multiple distinct regions of the prompt for each problem. We explore the impact of each component in an ablation study of our \newHE benchmark suite with Codex. We ran four versions of the \newHE prompts, with some or all regions translated:
\begin{itemize}
    \item \textbf{Original Prompt}: does not translate doctests or natural language terminology (e.g. prompts as in HumanEval);
    \item \textbf{Test-only Translation}: translates doctests but not Python-specific terminology;
    \item \textbf{Full Translation}: translates unit tests, doctests, and Python-terminology in the prompt; and
    \item \textbf{No Doctests}: removes doctests and does not translate natural language terminology.
\end{itemize}
For Codex's pass@1 results, translating doctests and Python-specific terminology has little impact on better-performing languages (\Cref{fig:ablation}). However, translating these components seems more important for certain languages: Bash, PHP, Perl, R, Rust, Swift, and TypeScript. 

We note that six of these languages are ones where Codex does not perform substantially better on \newMBPP than \newHE (\Cref{fig:total_results}). The performance degradation observed for these languages when doctests are removed from the \newHE
 problems suggests that the worse than expected performance on \newMBPP could be due to the lack of doctests in that benchmark suite.

 Overall, we find significant differences between the \textbf{Full Translation} and \textbf{Test-Only Translation} experiments ($p=0.03$), and between \textbf{No Doctests} and \textbf{Test-Only Translation} ($p<0.001$), but not between \textbf{No Translation} and \textbf{Test-Only Translation} ($p=0.2$). This suggests that the Python terminology translation has a small but reliable effect, and that the presence of the doctests is important, though their translation is not. 


\subsection{Type Annotations}\label{subsec:static-types}

\begin{figure*}[t]
    \centering
    \includegraphics[width=0.8\textwidth]{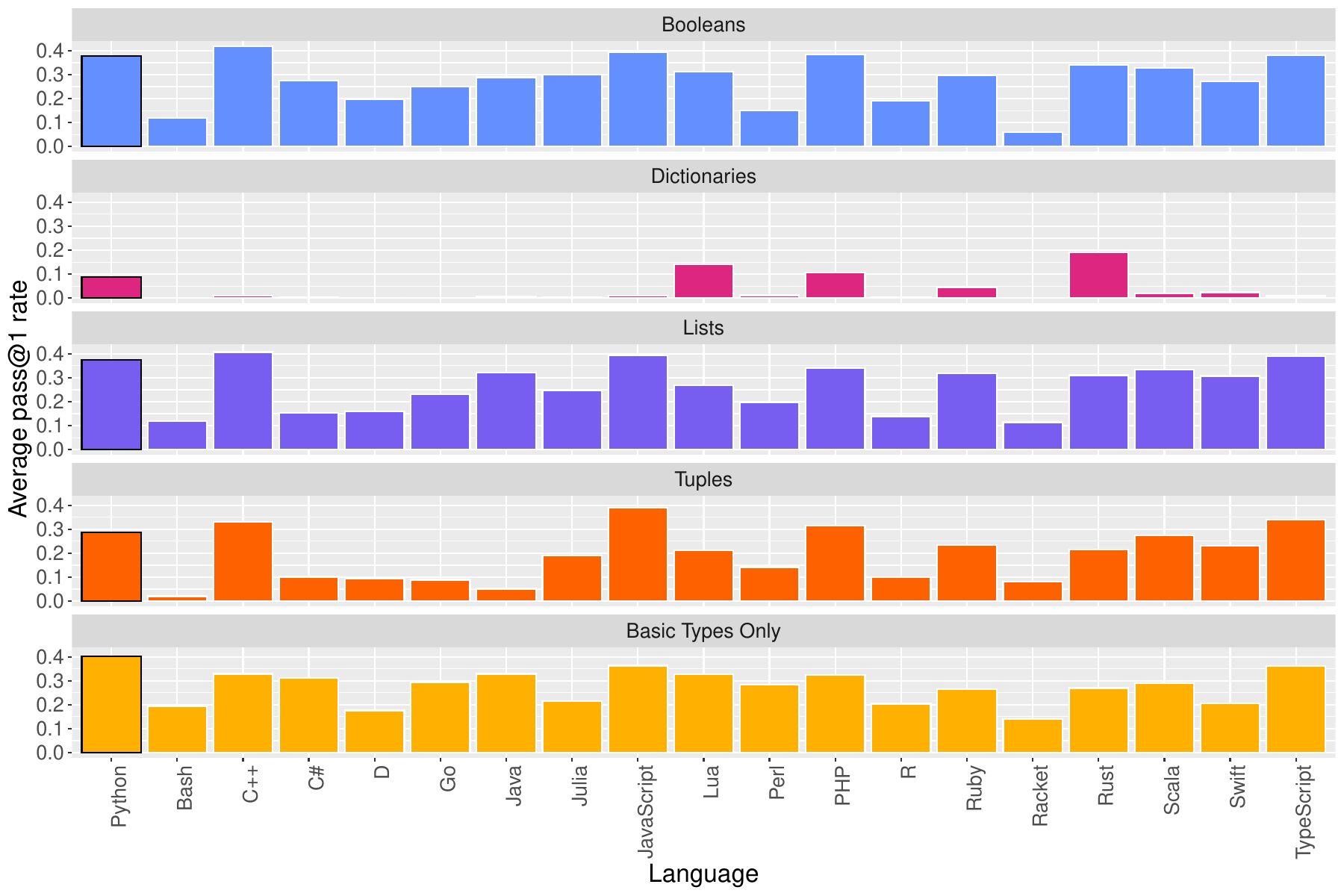}
    \caption{Impact of programming language features on Codex pass@1 performance by language}
    \label{fig:features}
\end{figure*}

One may conjecture that type annotations improve model performance by constraining the code generation search space. Or, perhaps, they might hurt performance by complicating the task, since the model must generate correct type annotations. 

In \cref{fig:he-lang-frequency} and \cref{fig:mbpp-lang-frequency}, the dashed lines in each category separate languages with type annotations (left) from languages without (right). We observe no overall effect of type annotations on Codex pass@1 rates on \newHE ($p=0.33$) or \newMBPP ($p=0.23$).

To explore the impact of type annotations at a more fine-grained level, we run a series of follow-up experiments using the \newHE benchmark suite. We focus on two languages: Python, which allows optional type annotations, and TypeScript, a gradually typed cousin of JavaScript. Gradual typing allows us to weaken type annotations or even configure the TypeScript compiler to ignore all type errors.

\subsubsection{Precise types improve TypeScript performance}
TypeScript has an ``Any'' type, which is compatible with all types. We run Codex on a variation of the \newHE TypeScript prompts where all types in the function signature are translated to ``Any''. We find that the loss of precise types hurts performance on TypeScript (-2.5\%; $p<0.001$).

\subsubsection{TypeScript type errors correlate with runtime errors}
Even gradual type-checking can reject programs that would in fact run without error. We run the Codex-generated TypeScript programs without first checking types. We observe no significant difference in pass@1 rates ($p=0.14$), suggesting that typed programs are rejected for genuine errors.

\subsubsection{Type annotations \emph{do not} affect Python performance}
We run a similar experiment with Codex and Python, where we remove all the type annotations from the \newHE prompts. We find that this has no significant effect on Codex's pass@1 rate for Python ($p=0.23$). 

\subsubsection*{} We interpret these results as evidence that type annotations do not guide search in general, since they do not improve Python performance, but that informative types are necessary for languages where type annotations are standard, perhaps in order to fit the token distribution of high-quality typed code seen in training. 

\subsection{Sensitivity to Compilation Choices}\label{subsec:prompt-engineering}

Each \system{} compiler makes small choices about how to translate prompts that could have an impact on performance. We explore some of these choices below.

\subsubsection{Comment style affects performance}
Most programming languages have several comment styles (e.g., single-line vs. multi-line). To investigate their impact, we ran follow-up experiments with Codex on the \newHE benchmark suite for two languages: PHP (\textsc{Medium}) and Racket (\textsc{Niche}). 

Our original prompts use single-line comments for both PHP and Racket, following conventional style. We re-ran Codex on versions of the \newHE problems for both languages using multi-line comments instead. This improves the pass@1 rate for Racket (+1.9\%, $p<0.001$), but decreases it for PHP (-3.1\% , $p=0.001$). 

\subsubsection{Naming arguments improves performance for Perl}

Functions in Perl do not have formal named arguments. Instead, all arguments are passed in a special array. Our compiler to Perl produces a prompt that pops elements off the special array and names them, with the expectation that this would improve model performance. 

We ran a follow-up experiment on a version of the \newHE problems for Perl where we omit this argument-naming prompt, so the model has to infer everything about arguments from the natural language description and examples. 
This significantly lowers Codex's pass@1 rate (-8\%; $p<0.001$).

\subsubsection*{}
In summary, our results show that code generation performance can be sensitive to prompt engineering choices for both high and low frequency languages. 

\subsection{Impact of Language Features}\label{subsec:features}

One challenge of extending existing benchmarks to new programming languages is that not all programming languages have the same features. Although the MBPP and HumanEval benchmarks consist of relatively simple functions, they exercise a variety of datatypes, not all of which have a straightforward equivalent in all programming languages in our dataset. 

To explore whether model success is impacted by the Python language features used in the program, we categorized all problems from the HumanEval benchmark suite into groups based on the Python language features used in their type annotations: Booleans, dictionaries, lists, tuples, or none of the above.

Figure \ref{fig:features} shows the performance by language on each type of problem. A mixed-effects model finds no significant effect of problem type, when programming language is treated as a random effect.

Many languages appear to struggle with questions involving tuples. Some of these are languages that lack a native tuple type, such as Java. However, JavaScript performs well despite lacking tuples. 

Although many languages show poor performance on dictionary problems, there are only 3 problems in this category, so these results should be interpreted with caution.

\subsection{Fine-grained Error Analysis}\label{subsec:errors} 
Code generation models generate many more failing programs---programs that produce errors or fail to pass unit tests---than programs which run successfully. This section presents a detailed evaluation of errors present in the Codex-generated completions for \newHE problems in 4 languages: Python, C\#, Swift, and Racket. See Appendix \ref{app:errors} for a full categorization.

We first identified specific error labels for each language and then grouped them into themes (e.g. ``NullReference''). We produced five general error categories: \textsc{Run-time}, \textsc{Static}, \textsc{Type}, \textsc{Language}, and \textsc{Model}. We group similar error sources together across languages, even if they occur in different contexts: for example, calling a function with a value of the wrong type may fail at compile-time or run-time, depending on the language's type system.

The most common \textsc{Static} theme across all languages is ``UndefinedIdentifier'', which contains errors related to referencing non-existent terms. These errors can be caused in many ways -- calls to functions not in the local context, 
use of Python-like keywords, 
or calls to methods from external libraries that were not imported.

Some errors in the \textsc{Runtime} category mimic those we expect from software engineers (e.g., index-out-of-range errors). However, others are unlike human mistakes. Notable themes in the latter group (\textsc{Model}) include generating code that throws exceptions on purpose and generating code in an entirely different language. For instance, Codex frequently generates Markdown code for Racket problems. Although we don't have access to the Codex dataset, we suspect that Racket is not well-represented in the dataset. We posit that these errors occur because Racket files begin with a language declaration (\lstinline|#lang racket|) that is easily mistaken for a Markdown heading.

Finally, the category \textsc{Language} includes multiple themes related to the specifics of the target language itself. The ``LanguageSpecific'' theme contains idiosyncratic errors such as the requirement of labeled arguments in Swift. ``DoesNotKnowSyntax'' includes errors in Racket caused by incorrectly generated core language constructs.

\section{Threats to Validity}

Our work translates two Python code generation benchmarks into \nlang{} other languages and evaluates the performance of three code generation models on the translated benchmarks.

The principal threat to validity is that the (translated) benchmarks may not be representative of the kinds of problems that programmers typically solve in each languages.
For example, we evaluate both scripting languages (e.g., Python and JavaScript) and systems languages (e.g., C++ and Rust) on the same task, but programmers frequently use these languages for very different tasks.
We characterize the HumanEval and MBPP benchmarks as a mix of basic programming problems and straightforward interview questions.
Thus, performance on benchmarks may not accurately represent real-world performance.

Code generation models are sensitive to small changes in how prompts are designed, as we show in our exploration of prompt design choices for three of our languages (\cref{subsec:prompt-engineering}).
It is likely that the pass rate on individual languages could be improved with even more language-specific effort.
We do provide an ablation study on prompts for all languages in our dataset (\cref{subsec:ablation}) to investigate the impact of our different translation components.

The quality of generated code is also sensitive to decisions about how to sample completions (\cref{subsec:sampling}).
We use the same sampling configuration that is used in most prior work on code generation. Empirical results show these settings are optimal for Python~\cite{chen2021evaluating}, but it is possible that different settings would be better for other programming languages.\footnote{This would be a very resource-intensive experiment, beyond the scope of an academic group. The original experiment on sampler configurations by Chen et al.~\cite{chen2021evaluating} has not been repeated by any lab.}
However, in a practical deployment of a multi-language code generation model, it may not be feasible to adjust the sampler for every input language.

\section{Related Work}

In this section we focus on related work on evaluating neural code generation models.

\newpara{Early approaches.}
Early work on neural network code generation relied on textual similarity metrics for evaluation. For instance, Feng et al. \cite{feng2020codebert} evaluate their CodeBERT model on six programming languages using BLEU~\citep{bleu}. Ren et al. \cite{ren2020codebleu} proposes a code generation-specific formulation of this metric. However, previous work has found that textual similarity metrics correlate only weakly with code correctness~\cite{ren2020codebleu,austin2021program,chen2021evaluating}, highlighting the importance of benchmark suites with unit tests.

\newpara{Other benchmark formats.}
The benchmarks that we translate pair code with comments; some other benchmarks pair natural language descriptions of other kinds. For instance, the CoNaLa~\citet{yin2018learning} benchmark consists of matching natural language questions and code snippets mined from StackOverflow. We note that MCoNaLA~\citet{wang2022mconala}, which extends CoNaLa to Spanish, Japanese, and Russian, is the only currently available benchmark for evaluating code generation from multiple natural languages.

\newpara{Other monolingual benchmarks.}
There are monolingual code generation benchmarks in languages beyond Python. 
Kulal et al. \citet{kulal2019spoc} presents a C++ dataset consisting of crowdsourced descriptions of lines of code. Iyer et al. \citet{iyer2018mapping} present a Java benchmark taken from online code repositories. Zhong et al. \citet{wikisql} and Yu et al. \citet{spidersql} propose benchmarks for SQL.
However, we do not believe SQL is amenable to translation, since conventional types in programming languages do not naturally translate to the types of relational tables. Moreover, of these datasets, only Kulal et al.'s includes unit tests to enable evaluation of code correctness \citet{kulal2019spoc}.
 
Our approach could be applied to other Python code generation benchmark suites like MathQA-Python~\citep{austin2021program}, a set of mathematical word problems with multiple choice answers, or APPS~\citep{hendrycks2021measuring}, a set of problems taken from open-access code competition websites like Codeforces.

\newpara{Other tasks.}
Although we focus specifically on benchmarks for the code generation task, there are many other tasks that have been used to evaluate code generation models, including generating unit tests from code~\citep{tufano2020unit}, code search~\citep{feng2020codebert,ahmed2022multilingual}, and type inference~\cite{wei:lambdanet,hellendoorn:dl-ti,pradel:typewriter}. Lu et al. \citet{lu2021codexglue} propose a suite of evaluation datasets for ten tasks, including code translation, docstring generation, and code summarization.

\newpara{Other code generation models.}
We evaluate three state-of-the-art code generation models, but many other models that have been proposed. Two influential early models were CodeBERT~\citep{feng2020codebert} and PyMT~\citet{clement-etal-2020-pymt5}. More recent models include PolyCoder~\citep{xu2022systematic}, CodeParrot \cite{tunstall2022natural}, AlphaCode~\citep{alphacode}, and PaLM-Coder~\citep{chowdhery2022palm}. PolyCoder was not trained on natural language text, and its authors explicitly state that it may not be suitable for NL2Code. AlphaCode and PaLM-Coder are not available for academic researchers to investigate.

\newpara{Other multi-language evaluation.}
Xu et al. \citet{xu2022systematic} measure the performance of several code generation models on 12 languages. However, they evaluate model performance using perplexity, rather than building a benchmark with unit tests, as we do; they test code correctness only for Python.

HumanEval-X\footnote{\url{http://keg.cs.tsinghua.edu.cn/codegeex}} is an unpublished benchmark that appeared after our work that manually translates the HumanEval problems into four languages (C++, Java, JavaScript, and Go).
Our compiler-based approach has the advantage of easy scalability: we support 18 languages and both HumanEval and MBPP.

MBXP~\citet{mbxp} is a concurrenteffort by Amazon Research to evaluate code generation models. We support more languages (13 vs. 19), though MBXP translates an additional benchmark (MathQA). Both MBXP and our work could be extended to support more languages and benchmarks.
However, there are deeper differences in the nature of our translation and evaluation:
\begin{itemize}

    \item We believe our approach to testing is more reliable. Rather than keeping the unit tests hidden from the model, MBXP prompts the model with the same unit tests it uses to test the generated code. Thus the code generation model can ``see'' the test cases that it will be evaluated on. In contrast, we use a hidden set of unit tests to evaluate code correctness.

    \item Our work is more faithful in translating types from Python into typed languages. For example, our type inference
    infers types like \lstinline|Either[X,Y]| and \lstinline|Optional[X]| and translates them to algebraic datatypes in typed languages (\Cref{subsec:static-types}). MBXP produces types such as \lstinline|Object| and \lstinline|Any| in languages like Java and Scala, which are less idiomatic.
    For languages that do not support \lstinline|Any|, such as C++, MBXP fail to translate these benchmarks altogether.

    \item MBXP uses \emph{greedy decoding} in their evaluation of public models. Greedy decoding produces a single candidate program which may not be the most likely program. 
    Prior work has shown that sampling the output of a code generation model significantly improves the correctness of generated code~\cite{chen2021evaluating}. We follow standard practices for sampling (\Cref{subsec:sampling}).

\end{itemize}

\section{Conclusion}

We propose \system, the first massively parallel, multi-language benchmark for natural-language-to-code generation. We write compilers to translate code generation benchmarks from Python to \nlang{} additional programming languages that span a spectrum of language features and popularity.

We translate two widely used unit test-driven benchmarks for code generation: HumanEval and MBPP. Using our multi-language parallel versions, we present the first multi-language code correctness evaluation of three state-of-the-art models: Codex, CodeGen, and InCoder. We demonstrate that Codex displays high performance across a variety of programming languages, performing similarly to Python on several languages, most notably, JavaScript. 

In our detailed by-language analysis, we find a predictable effect of language frequency, but draw mixed conclusions about the impact of type annotations. Our detailed error analysis highlights common patterns in four languages, finding model errors that are both like and unlike those of human programmers. We hope that our in-depth, parallel evaluation of a large set of languages will be a useful guide for developers weighing whether the utility of code generation tools in their project context.

Our publicly available benchmark is also easy to extend to new  problems and languages. We hope it will help evaluate and develop future work on multi-language code generation models.

\ifanon

\else

\section{Acknowledgments}

We thank Steven Holtzen and Joydeep Biswas for loaning us their GPUs. We thank Northeastern Research Computing for technical support, especially Greg Shlomo. This work was partially supported by the National Science Foundation grant CCF-2052696.

\fi


\bibliographystyle{IEEEtranN}
\bibliography{paper.bib,arjun_multipleval.bib}

\newpage
\appendices
\onecolumn

\section{Details of Language Translations}\label{app:lang-eval}

The tables below describe the details of all \nlang language translations as well as our Python translation. Technical information regarding running experiments and evaluating generated programs can be found at \url{github.com/nuprl/MultiPL-E}. Here we address language-specific decisions that are relevant to the prompt translation task. Specifically, we outline the following details: 
\begin{enumerate}
    \item The language version used as a reference for creating the value-to-value translation. 
    \item The stop tokens used for signaling the end of program generation. Across languages these reflect terms that begin and/or start code blocks (variations of \texttt{\textbackslash n\}} are common).
    \item Details about prompt creation. This sections highlight the choice of comments and any necessary preamble information (e.g., the opening tag \texttt{<?php} in PHP)
    \item Information about mapping values and/or types from Python. Notes here describe places where case-by-case decisions need to be made, or a language's limitations required not converting a subset of values and/or types. Omitted discussions represents straightforward conversions (e.g., integers in Perl). 

\end{enumerate}

\begin{table}[H]
    \centering
    \begin{tabularx}{\textwidth}{|g|| X|}\hline
    \cellcolor{red!25}\textsc{Bash} & \\\hline\hline
    Reference Version for Translation & 5.1.16\\\hline
    Stop Tokens & '\texttt{\textbackslash n}\}' \\\hline
    Prompt Information & We translate each Python docstring to Bash comments (each line prefixed with \texttt{\#}). Each Python function signature is translated to a Bash function signature, which is of the form \texttt{function\_name()}, as Bash functions do not have explicit parameters. Type annotations are translated to comments in the prompt, which describe the encoding (or none) for each of the function parameters. The shebang (\texttt{\#!/bin/bash}) was prepended to the prompt. \\ \hline

    Type Translations & Bash is not a general-purpose programming language, and its many quirks make translation challenging, particularly for data structures like lists and maps. While Bash has numerically indexed and string-associative arrays, the shell's ecosystem typically works with these structures in string-y formats: lists are typically whitespace separated elements; associative maps are in formats like comma-separated values (CSV). We use those conventions in our type translations. \\\hline
    \end{tabularx}
    \label{tab:bash_info}
\end{table}

\begin{table}[H]
    \centering
    \begin{tabularx}{\textwidth}{|g|| X|}\hline
    \cellcolor{red!25}\textsc{C++} & \\\hline\hline
    Reference Version for Translation & C++17 compiled using g++17\\\hline
    Stop Tokens & '\texttt{\textbackslash n}\}' \\\hline
    Prompt Information & Each prompt contains C++ single line comments where each line is prefixed with \texttt{//}. Python function signatures are translated to C++ signatures and we add \texttt{\#include} statements. \\ \hline
    Type Translations & All Python integers are translated to C++ \texttt{long} and Python floats are translated to C++ \texttt{float}. A Python list is translated to \texttt{std::vector}, a dictionary to \texttt{std::map}, a tuple to \texttt{std::tuple}, a string to \texttt{std::string}, and Python's \texttt{Any} type to \texttt{std::any}. A new C++ union type is declared for each union type annotation in Python. \\\hline
    \end{tabularx}
    \label{tab:cpp_info}
\end{table}

\begin{table}[H]
    \centering
    \begin{tabularx}{\textwidth}{|g|| X|}\hline
    \cellcolor{red!25}\textsc{C\#} & \\\hline\hline
    Reference Version for Translation & C\# 5 with Mono 6.12\\\hline
    Stop Tokens & '\texttt{\textbackslash n ~~~~~\}\textbackslash n'}\\\hline
    Prompt Information & The prompt contains a class declaration with the translated method as its \texttt{public static} member and C\# single line comments, where each line is prefixed with \texttt{//}. Adding a member of class also adds indentation to each line inside class declaration (note the indentation in the stop token). All function and argument names are converted to C\#'s naming convention where the first letter of all words is in capital case.\\ \hline
    Type Translations & Most types were translated to their C\# direct equivalent (e.g. Python \texttt{tuple} to C\# \texttt{tuple}). There are some exceptions: Python \texttt{int} is translated to a C\# \texttt{long} and Python’s \texttt{Any} type annotation is translated to C\# object. Since C\# does not support union types, we do not convert Python union annotations. \\\hline
    \end{tabularx}
    \label{tab:csharp_info}
\end{table}

\begin{table}[H]
    \centering
    \begin{tabularx}{\textwidth}{|g|| X|}\hline
    \cellcolor{red!25}\textsc{D} & \\\hline\hline
  Reference Version for Translation & dmd 2.100.0 \\\hline
    Stop Tokens & \texttt{'\textbackslash n\textbackslash n', '\textbackslash void', '\textbackslash bool', '\textbackslash int'} \\\hline
    Prompt Information & The prompt was given as a multi-line comment (\texttt{/* ... */}). \\ \hline
  
    
    
    
    Type Translations & 
        Most types in Python have equivalents in D. One exception is Python integers, which we translate to \texttt{long}. 
        Dictionaries are translated to \texttt{Nullable} of associative arrays, a built-in array that supports indices of any types. 
        Associative arrays must be non-empty in D, so the \texttt{Nullable!(...)} template type is needed to wrap around the associative array,
        i.e. an empty array is denoted as the ``null'' state.
        Tuples are translated to the \texttt{Tuples!(...)} template type; however, the tuple type in D cannot be variable arity. 
        Union types and \texttt{Any} are not translated.
    \\\hline
    
    \end{tabularx}
    \label{tab:d_info}
\end{table}

\begin{table}[H]
    \centering
    \begin{tabularx}{\textwidth}{|g|| X|}\hline
    \cellcolor{red!25}\textsc{Go} & \\\hline\hline
    Reference Version for Translation & 1.18.1\\\hline
    Stop Tokens & \verb|'\nfunc', 'struct', '\n// '| \\\hline
    Prompt Information & The prompt is translated as a line comment (with \texttt{//}) above the function stub. For short functions, it is recommended to use single line comments. \\ \hline

    Type Translations & Python Lists and Dictionaries were mapped to Go's Slices and Maps, respectively. Since Go requires type annotations, we utilized Python's type annotations to both translate the candidate function and the tests. Go requires explicitly declaring types for a compound datatype (e.g., a Python list \texttt{[1, 2, 3]} translates to \texttt{[]int\{1, 2, 3\}}). Go does not have an equivalent \texttt{Union}, \texttt{Option}, or \texttt{Tuple} data type, but it is possible to create a non-homogenous slice using \texttt{[]interface{}} -- therefore we reject the two former and we convert the latter.\\\hline
    Other Notes & We consulted the following style guide as part of our translation to Go (\url{https://go.dev/doc/effective_go}).\\\hline
    \end{tabularx}
    \label{tab:go_info}
\end{table}

\begin{table}[H]
    \centering
    \begin{tabularx}{\textwidth}{|g|| X|}\hline
    \cellcolor{red!25}\textsc{Java} & \\\hline\hline
    Reference Version for Translation & OpenJDK 17\\\hline
    Stop Tokens & '\texttt{\textbackslash n ~~~~      
       \}\textbackslash n'}\\\hline
    Prompt Information & The prompt contains a class declaration with the translated method as its \texttt{public static} member and Java single line comments, where each line is prefixed with \texttt{//}. Adding a member of class also adds indentation to each line inside class declaration (note in the intention in the stop token). All function and arguments are converted to Java's naming convention where the first letter is lowercase and the first letter of all other words are capitalized. \\ \hline
    Type Translations & The type translation from Python to Java is performed by translating a Python \texttt{int} to a Java \texttt{long}, Python \texttt{float} to Java \texttt{float}, a Python \texttt{list} to \texttt{Vector}, a \texttt{dictionary} to \texttt{HashMap}, a string to \texttt{String}, and Python’s \texttt{Any} type annotation to Object. Since OpenJDK does not support tuples, we use \texttt{javatuples} library and translates Python tuples to \texttt{javatuples.Tuple}. Since Java does not support union types, we do not convert Python union annotations. \\\hline
    \end{tabularx}
    \label{tab:java_info}
\end{table}

\begin{table}[H]
    \centering
    \begin{tabularx}{\textwidth}{|g|| X|}\hline
    \cellcolor{red!25}\textsc{JavaScript} & \\\hline\hline
    Reference Version for Translation & 18.6\\\hline
    Stop Tokens & \verb|'\nfunction ', '\n/*', '\n//', '\nconsole.log'| \\\hline
    Prompt Information &  We convert the Python prompt into a block of comments using \texttt{//}.\\ \hline
    
    Type Translations & Most type translations are direct. Python lists and tuples were translated into JS arrays. Dictionaries were translated into objects.\\\hline\hline
    \end{tabularx}
    \label{tab:js_info}
\end{table}

\begin{table}[H]
    \centering
    \begin{tabularx}{\textwidth}{|g|| X|}\hline
    \cellcolor{red!25}\textsc{Julia} & \\\hline\hline
    Reference Version for Translation & 1.7.3 \\\hline
    Stop Tokens & \verb|'\nfunction', '\nmacro', '\n\n'| \\\hline

    Prompt Information & Julia shares both its documentation and line comment
        syntax with Python, and thus the prompt is left unchanged by the
        translation.\\\hline





    Type Translations & We translate Python’s \texttt{int} to \texttt{Int64},  \texttt{float} to
        \texttt{Float64}, and \texttt{List} to \texttt{Vector}. The only coercion required in the
        benchmarks come from the fact that Julia generates the type \texttt{Vector\{Any\}}
        for the unannotated empty vector. Thus, if the empty vector is given as an argument to the function, it is coerced to the expected (more
        specific) type. Julia has first-class support for \texttt{Union} types;
        therefore, we represent Unions directly and \texttt{Optional<T>} as the type
        \texttt{Union\{T, Nothing\}}. \\\hline
    \end{tabularx}
    \label{tab:julia_info}
\end{table}

\begin{table}[H]
    \centering
    \begin{tabularx}{\textwidth}{|g|| X|}\hline
    \cellcolor{red!25}\textsc{Lua} & \\\hline\hline
    Reference Version for Translation & 5.3\\\hline
    Stop Tokens & \verb|'\nlocal', '\nfunction', '\n--', '\n\n'| \\\hline
    Prompt Information & We convert the Python prompt to a block of single-line comments using \texttt{---}. \\ \hline
    Type Translations & The only data structure in Lua is a table, and tables with integer indices behave like lists. Thus we translate Python dictionaries, tuples, and lists to tables. \\\hline
    \end{tabularx}
    \label{tab:lua_info}
\end{table}

\begin{table}[H]
    \centering
    \begin{tabularx}{\textwidth}{|g|| X|}\hline
    \cellcolor{red!25}\textsc{Perl} & \\\hline\hline
    Reference Version for Translation & 5.34 \\\hline
    Stop Tokens & \texttt{'\textbackslash nsub', '\textbackslash n\#', '\textbackslash n\textbackslash n'} \\\hline
    Prompt Information & We convert the Python prompt to a block of single-line comments, using \texttt{\#}.  \\ \hline
    
    Type Translations & We are careful to pass data structures by reference; we translate Python lists and tuples to anonymous arrays, and dictionaries to anonymous hashes. Perl lacks a Boolean type; we translate \texttt{True} to 1 and \texttt{False} to the empty string, since these are the values returned by logical operators. \\\hline
    \end{tabularx}
    \label{tab:perl_info}
\end{table}

\begin{table}[H]
    \centering
    \begin{tabularx}{\textwidth}{|g|| X|}\hline
    \cellcolor{red!25}\textsc{PHP} & \\\hline\hline
    Reference Version for Translation & 8.1.2 (cli)\\\hline
    Stop Tokens & \texttt{'\textbackslash nfunction', '\textbackslash n?>', '\textbackslash n\/\/', '\textbackslash n\#'}\\\hline
    Prompt Information & In our full translation, the prompt was given as single-line comments, using \texttt{//}, rather than using PHP's two other comment styles (single line \texttt{\#} and multi-line \texttt{/* ... */}). The PHP opening tag, \texttt{<?php}, was prepended to the prompt, and the closing tag was omitted, following the recommendation for a file that only contains PHP code (\url{https://www.php.net/manual/en/language.basic-syntax.phptags.php}). \\ \hline
    
    Type Translations & PHP arrays are actually ordered maps, so Python lists, tuples, and dictionaries were translated to arrays. Arrays were defined using the default syntax, \texttt{array()}, instead of the shorthand \texttt{[]}. Strings are double quoted, and Python's \texttt{None} is translated to \texttt{null}. \\\hline
    \end{tabularx}
    \label{tab:php_info}
\end{table}

\begin{table}[H]
    \centering
    \begin{tabularx}{\textwidth}{|g|| X|}\hline
    \cellcolor{red!25}\textsc{Python} & \\\hline\hline
    Reference Version for Translation & 3.10 \\\hline
    Stop Tokens &  \texttt{'\textbackslash ndef', '\textbackslash n\#', '\textbackslash nif', '\textbackslash nclass'}\\\hline
    Prompt Information &  The prompt was presented as in the original HumanEval dataset: a multi-line docstring. If type annotations were present, the typing library was imported via an import statement at the beginning of the prompt.\\ \hline
    Type Translations & The Python translation is trivial: each type is translated to itself.\\\hline\hline
    \end{tabularx}
    \label{tab:py_info}
\end{table}

\begin{table}[H]
    \centering
    \begin{tabularx}{\textwidth}{|g|| X|}\hline
    \cellcolor{red!25}\textsc{R} & \\\hline\hline
    Reference Version for Translation & 4.1\\\hline
    
    Stop Tokens & \texttt{'\textbackslash n\#','\textbackslash n```'} \\\hline
    Prompt Information & We convert the Python prompt to a block of single-line comments using \texttt{\#}. \\ \hline
    Type Translations & R vectors are more commonly used than R lists; however, R vectors are restricted to storing homogenous data types. We translate Python Lists and Tuples to R vectors using the \texttt{c()} function when possible (i.e., when the contents are homogenous), and to R lists using the \texttt{list()} function otherwise. We convert Python dictionaries to named lists. R, like Python, supports both single and double quoted strings. \\\hline 
    \end{tabularx}
    \label{tab:r_info}
\end{table}

\begin{table}[H]
    \centering
    \begin{tabularx}{\textwidth}{|g|| X|}\hline
    \cellcolor{red!25}\textsc{Racket} & \\\hline\hline
    Reference Version for Translation & 8.2\\\hline
    Stop Tokens &  \texttt{'\textbackslash n(define ', '\textbackslash n\#|', '\textbackslash n;', '\textbackslash n('}\\ \hline
    Prompt Information & We convert the Python prompt to a block of single-line comments using ';'. \\ \hline
    Type Translations & We translate Python Lists and Tuples to Racket lists using \texttt{(list )}. We convert Python dictionaries to hash maps using \texttt{(hash )}. Racket does not support single-quoted strings, so we convert all strings to double-quoted strings.\\\hline
    \end{tabularx}
    \label{tab:racket_info}
\end{table}

\begin{table}[H]
    \centering
    \begin{tabularx}{\textwidth}{|g|| X|}\hline
    \cellcolor{red!25}\textsc{Ruby} & \\\hline\hline
    Reference Version for Translation & 3.0.2\\\hline
    Stop Tokens & \texttt{'\textbackslash nclass', '\textbackslash ndef', '\textbackslash n\#', '\textbackslash n\textbackslash n'}\\\hline
    Prompt Information & Although there are block comments in Ruby (\texttt{=begin ... =end}), they are discouraged by community style guides. Therefore, the prompt was converted to a block of single-line comments prefixed by \texttt{\#}. \\ \hline
    Type Translations & Python Lists and Tuples were mapped to Ruby Arrays with the \texttt{[...]} shorthand per style guides. The idiomatic \texttt{=>} Ruby syntax was used for dictionary creation. While Ruby supports both double- and single-quoted strings, Python strings were converted to double-quoted Ruby strings as they work with string interpolation. \\\hline\hline
    Other Notes & We consulted the following two style guides as part of our translation to Ruby (\url{https://ruby-style-guide.shopify.dev/}, \url{https://github.com/rubocop/ruby-style-guide}).  \\\hline
    \end{tabularx}
    \label{tab:ruby_info}
\end{table}

\begin{table}[H]
    \centering
    \begin{tabularx}{\textwidth}{|g|| X|}\hline
    \cellcolor{red!25}\textsc{Rust} & \\\hline\hline
    Reference Version for Translation & 1.59.0\\\hline
    Stop Tokens & '\texttt{\textbackslash n}\}' \\\hline
    Prompt Information & A doc comment is used to indicate that the prompt
    information corresponds to the behavior of the function and not internal
    implementation details (each line prefixed with \texttt{///}). No arguments
    are annotated with \texttt{mut} - in all cases (we used owned values) they
    can be moved to a mutable variable if necessary, and unnecessary mutable
    annotations may be confusing.
     \\ \hline
    Type Translations & All annotated values are owned. While in Rust
    it sometimes makes sense to accept borrowed values (for example, if no
    mutation or move is necessary), it is difficult to infer when this is
    appropriate from the Python signature or prompt. Inferring when a
    borrowed result type could be used would be even more difficult. Thus, \texttt{str}
    is translated as \texttt{String} and \texttt{List} is translated to
    \texttt{Vec}. \texttt{Tuple} is translated
    to Rust's tuple, \texttt{dict} to Rust's \texttt{std::collections::HashMap}, and
    \texttt{Optional} to \texttt{Option}. While Python's
    \texttt{int} must support at least 64 bit integers, the more idiomatic
    \texttt{isize} is used to represent them in Rust. Python's \texttt{float}
    is translated to the corresponding \texttt{f64} and \texttt{bool} to
    \texttt{bool}. Problems annotated with a \texttt{Union}, \texttt{Any}, or
    \texttt{Ellipsis} are not supported. \\\hline
    \end{tabularx}
    \label{tab:rust_info}
\end{table}

\begin{table}[H]
    \centering
    \begin{tabularx}{\textwidth}{|g|| X|}\hline
    \cellcolor{red!25}\textsc{Scala} & \\\hline\hline
    Reference Version for Translation & Scala 2.23\\\hline
    Stop Tokens & '\texttt{\textbackslash n~~~~\}\textbackslash n}' \\\hline
    Prompt Information & The prompt contains a class declaration with the translated method as its member and Scala single line comments where each line is prefixed with \texttt{//}. Adding a member of class also adds indentation to each line inside class declaration (note the indentation in the stop token).
    All function and argument names are converted Scala's naming convention where the first letter is lowercase and the first letter of all other words is in capital case. \\ \hline
    Type Translations & The type translation from Python to Scala is performed by translating a Python \texttt{int} to a Scala \texttt{long}, Python \texttt{float} to Scala \texttt{float}, a Python \texttt{list} to Scala \texttt{List}, a Python \texttt{dictionary} to Scala \texttt{Dictionary}, a Python string to Scala string, a Python \texttt{tuple} to Scala \texttt{Tuple}, and Python’s \texttt{Any} type annotation to Scala \texttt{Any}. Python union annotations of two types is converted to Scala's \texttt{Either} type. Problems with \texttt{Union} of more than two types are not supported.\\\hline
    \end{tabularx}
    \label{tab:scala_info}
\end{table}

\begin{table}[H]
    \centering
    \begin{tabularx}{\textwidth}{|g|| X|}\hline
    \cellcolor{red!25}\textsc{Swift} & \\\hline\hline
    Reference Version for Translation & 5.8\\\hline
    Stop Tokens & '\texttt{\textbackslash n}\}' \\\hline
    Prompt Information & The prompt is given by doc comments (prepended with \texttt{///}). For documenting function behavior, doc comments are preferred over standard comments. \\ \hline
    Type Translations & Python Lists, Dictionaries and Tuples were mapped to Swift Lists, Dictionaries and Tuples, respectively. Untyped Python parameters
    were mapped to \texttt{AnyHashable} in Swift, as opposed to \texttt{Any}, as it allows for equality comparisons and storage in dictionaries, so is the closest
    equivalent to untyped Python values. Optional types or Unions with \texttt{None} in Python were converted to \texttt{?} optional types in Swift, binary Union types were converted to \texttt{Result} types, and larger Union types were converted to generated algebraic datatypes. The generated algebraic datatype definitions (and \texttt{Error} protocol conformance in the case of \texttt{Result}) were inserted into the prompt, above the doc comments.\\\hline\hline
    Other Notes & We consulted the following style guide as part of our translation to Swift (\url{https://www.swift.org/documentation/api-design-guidelines/}).\\\hline
    \end{tabularx}
    \label{tab:swift_info}
\end{table}

\begin{table}[H]
    \centering
    \begin{tabularx}{\textwidth}{|g|| X|}\hline
    \cellcolor{red!25}\textsc{TypeScript} & \\\hline\hline
    Reference Version for Translation & TypeScript compiler version 4.5, Node version 18.6\\\hline
    Stop Tokens & \verb|'\nfunction ', '\n/*', '\n//', '\nclass'| \\\hline
    Prompt Information &  We convert the Python prompt into a block of comments using \texttt{//}.\\ \hline
    Type Translations & Types are translated by utilizing the annotations provided in our Python tests. Lists and tuples were translated into arrays. Dictionaries were translated into objects.\\\hline
    \end{tabularx}
    \label{tab:ts_info}
\end{table}

\section{Datasheet}

The datasheet below follows the categories proposed in \cite{gebru2021datasheets}:
\begin{quote}
Timnit Gebru, Jamie Morgenstern, Briana Vecchione, Jennifer Wortman Vaughan, Hanna Wallach, Hal Daumé III, and Kate Crawford. "Datasheets for datasets." Communications of the ACM, 2021, 86-92.
\end{quote}

\subsection{Motivation}

\begin{itemize}

\item \textbf{For what purpose was the dataset created?}

The datasets were originally created to evaluate the performance of code generation models. They were translated from Python to other programming languages to extend evaluation to new languages. 

\item \textbf{Who created the dataset?}

HumanEval was originally created by \citet{chen2021evaluating}. MBPP was originally created by \citet{austin2021program}. Both datasets were modified by the authors of this paper.

\item \textbf{Who funded the creation of the dataset?}
\ifanon
\emph{Omitted for review.}
\else
This work was partially supported by the National Science Foundation.
\fi

\end{itemize}

\subsection{Composition}

\begin{itemize}

\item \textbf{What do the instances that comprise the dataset represent?}

The instances of the dataset represent programming problems in 18 programming languages.

\item \textbf{How many instances are there in total?}

For \newHE, there are 3,059 instances (the modified set of 161 Python problems multiplied by 19 programming languages).

For \newMBPP, there are 7,619 instances (the modified set of 401 Python problems multiplied by 19 programming languages).

\item \textbf{Does the dataset contain all possible instances or is it a sample (not necessarily random) of instances from a larger set?}

The \newHE and \newMBPP are cleaned versions of the original dataset as described in \cref{limitations}. The \newHE dataset excludes 3 of the 164 original problems.

\item \textbf{What data does each instance consist of?}
Each instance is a programming problem with a problem description in natural language, a function signature, and unit tests.

\item \textbf{Is there a label or target associated with each instance?}

Each instance is numbered and labeled by the name of the function it tests and the language it is written in.

\item \textbf{Is the dataset self-contained, or does it link to or otherwise rely on external resources?}

The dataset is self-contained.

\end{itemize}

\subsection{Collection process}

\begin{itemize}

\item \textbf{How was the data associated with each instance acquired?}

The original Python datasets were manually cleaned. The versions for other programming languages and prompt variations were produced by a suite of compilers.




\item \textbf{Over what timeframe was the data collected?} 

May--October 2022

\item \textbf{Were any ethical review processes conducted?}

Not applicable. The dataset adapts a open source dataset released under the terms of the MIT license.

\end{itemize}

\subsection{Preprocessing/cleaning/labeling}

\begin{itemize}

\item \textbf{Was any preprocessing/cleaning/labeling of the data done?}

We added missing type annotations, formatted examples to use docstrings consistently, and changed random tests into unit tests in two problems.

\item \textbf{Was the "raw" data saved in addition to the preprocessed/cleaned/labeled data?}

The original data is for both datasets is publicly available.

\item {Is the software that was used to preprocess/clean/label the data available?}

The cleaning process described above was manual.

\end{itemize}

\subsection{Uses}

\begin{itemize}

\item \textbf{Has the dataset been used for any tasks already?}

The dataset has been used for evaluating the performance of code generation models.

\item \textbf{Is there a repository that links to any or all papers or systems that use the dataset?}

\ifanon
\emph{Omitted for review}
\else
\url{https://github.com/nuprl/MultiPL-E}
\fi

\item \textbf{What other tasks could the dataset be used for?} 

The dataset could be used to evaluate other LLMs of code, or potentially to improve their performance.



\end{itemize}

\subsection{Distribution}

\begin{itemize}

\item \textbf{Will the dataset be distributed to third parties outside of the entity? }

Yes. 

\item \textbf{How will the dataset be distributed?}

The dataset is publicly available at
\ifanon
\emph{Omitted for review}
\else
\url{https://github.com/nuprl/MultiPL-E}
\fi

\item \textbf{When will the dataset be distributed?}

Immediately.

\item \textbf{Will the dataset be distributed under a copyright or other intellectual property (IP) license, and/or under applicable terms of use (ToU)?}

No.

\item \textbf{Have any third parties imposed IP-based or other restrictions on the data associated with the instances?}

No.

\item \textbf{Do any export controls or other regulatory restrictions apply to the dataset or to individual instances?}

No.

\end{itemize}

\subsection{Maintenance}

\begin{itemize}

\item \textbf{Who will be supporting/hosting/maintaining the dataset?}

The original authors.

\item \textbf{How can the owner/curator/manager of the dataset be contacted?}

See the dataset website.

\item \textbf{Is there an erratum?}

No. Any identified and confirmed errors will be acknowledged as part of the repository. 

\item \textbf{Will the dataset be updated (for example, to correct labeling errors, add new instances, delete instances)?}

Yes.

\item \textbf{Will older versions of the dataset continue to be supported/hosted/maintained?}

Yes.

\item \textbf{If others want to extend/augment/build on/contribute to the dataset, is there a mechanism for them to do so?}

Yes, as described in the paper and website.

\end{itemize}

\section{Complete Statistical Findings}\label{app:mem}

We use binomial mixed-effects models fitted with the lme4 library in R for significance testing. A binomial distribution is appropriate because our outcomes consist of proportions of successes and failures; we use the number of completions (200, except in rare failure cases) as weights. 

We fit models to the Codex pass@1 completion rates in all experiments reported below. We treat problem number as a random effect to account for variability inherent to per-problem differences. For comparisons that do not break down effects by language, we also include language as a random effect. We include random slopes and intercepts for random effects except where noted.

Values that are statistically significant with a threshold of $p=0.5$ are displayed in \textbf{bold}.

\subsection{\newHE Mixed-Effects Results from \Cref{subsec:newhe-results}} 

To quantify the differences in performance among programming languages, a model with a fixed effect of programming language and random effects for problem number was fitted to the \newHE Codex pass@1 data.

Dummy coding was used with Python as the reference level; slopes for each language indicate differences between the pass@1 rate for Python and that language.

Table \ref{tab:mem_he_lang_codegen} shows the full estimates found by the model.

\begin{table}[H]
    \centering
    \begin{tabular}{|llll|}\hline
Fixed effects&$\widehat{\beta}$&$z$&$p$\\\hline
Intercept&-0.48 (+/- 0.4)&-1.1&0.27\\
Bash&-2.59 (+/- 0.3)&-7.7&< \textbf{0.0001}\\
C++&0.10 (+/- 0.4)&0.3&0.77\\    
C\#&-4.09 (+/- 0.6)&-7.2&< \textbf{0.0001}\\
D&-4.79 (+/- 0.5)&-9.7&< \textbf{0.0001}\\
Go&-2.61 (+/- 0.4)&-6.5&< \textbf{0.0001}\\
Java&-1.28 (+/- 0.3)&-3.9&< \textbf{0.0001}\\
Julia&-1.91 (+/- 0.4)&-5.2&< \textbf{0.0001}\\
JavaScript&-0.27 (+/- 0.3)&-0.8&0.43\\    
Lua&-1.04 (+/- 0.4)&-2.8&\textbf{0.005}\\ 
Perl&-2.0 (+/- 0.4)&-5.3&< \textbf{0.0001}\\
PHP&-0.30 (+/- 0.4)&-0.8&0.40\\  
R&-3.69 (+/- 0.4)&-8.5&< \textbf{0.0001}\\
Ruby&-0.68 (+/- 0.3)&-2.3&\textbf{0.024}\\ 
Racket&-3.78 (+/- 0.4)&-9.8&< \textbf{0.0001}\\
Rust&-1.07 (+/- 0.3)&-3.4&< \textbf{0.0001}\\
Scala&-0.52 (+/- 0.3)&-1.6&0.10\\   
Swift&-1.8 (+/- 0.3)&-5.7&< \textbf{0.0001}\\
TypeScript&-0.27 (+/- 0.3)&-0.9&0.39\\\hline
    \end{tabular}
    \caption{Mixed-effects results for Codex \newHE language comparison}\label{tab:mem_he_lang_codex}
\end{table}

A similar model was fit to the CodeGen pass@1 data, but without random slopes, because the very low pass rates for many problems makes the random effects estimates unstable. Table \ref{tab:mem_he_lang_codegen} shows the full estimates found by the model.

\begin{table}[H]
    \centering
    \begin{tabular}{|llll|}\hline
Fixed effects&$\widehat{\beta}$&$z$&$p$\\\hline
Intercept&-3.38 (+/- 0.2)&-18.5&< \textbf{0.0001}\\
Bash&-5.66 (+/- 0.04)&-145.8&< \textbf{0.0001}\\
C++&-0.21 (+/- 0.01)&-14.6&< \textbf{0.0001}\\    
C\#&-1.74 (+/- 0.02)&-112.7&< \textbf{0.0001}\\
D&-1.72 (+/- 0.02)&-111.6&< \textbf{0.0001}\\
Go&-1.0 (+/- 0.01)&-66.7&< \textbf{0.0001}\\
Java&-0.14 (+/- 0.01)&-10.2&< \textbf{0.0001}\\
Julia&-9.41 (+/- 0.2)&-44.0&< \textbf{0.0001}\\
JavaScript&0.12 (+/- 0.01)&8.6&< \textbf{0.0001}\\    
Lua&-2.43 (+/- 0.02)&-144.8&< \textbf{0.0001}\\ 
Perl&-3.0 (+/- 0.02)&-161.9&< \textbf{0.0001}\\
PHP&-1.74 (+/- 0.02)&-112.9&< \textbf{0.0001}\\  
R&-2.42 (+/- 0.02)&-114.2&< \textbf{0.0001}\\
Ruby&-22.40 (+/- 140)&-0.2&0.87\\ 
Racket&-5.51 (+/- 0.04)&-150.4&< \textbf{0.0001}\\
Rust&-3.78 (+/- 0.02)&-173.6&< \textbf{0.0001}\\
Scala&-3.65 (+/- 0.02)&-172.9&< \textbf{0.0001}\\   
Swift&-4.08 (+/- 0.02)&-174.0&< \textbf{0.0001}\\
TypeScript&-0.04 (+/- 0.01)&-2.7&< \textbf{0.0001}\\\hline
    \end{tabular}
    \caption{Mixed-effects results for CodeGen \newHE language comparison}\label{tab:mem_he_lang_codegen}
\end{table}

A similar model was fit to the InCoder pass@1 data, but without random slopes, because the very low pass rates for many problems makes the random effects estimates unstable. Table \ref{tab:mem_he_lang_incoder} shows the full estimates found by the model.

\begin{table}[H]
    \centering
    \begin{tabular}{|llll|}\hline
Fixed effects&$\widehat{\beta}$&$z$&$p$\\\hline
Intercept&-4.35 (+/- 0.3)&-14.5&< \textbf{0.0001}\\
Bash&-3.3 (+/- 0.03)&-131.1&< \textbf{0.0001}\\
C++&-0.99 (+/- 0.02)&-57.1&< \textbf{0.0001}\\
C\#&-1.94 (+/- 0.02)&-100.9&< \textbf{0.0001}\\
D&-3.97 (+/- 0.03)&-139.5&< \textbf{0.0001}\\
Go&-1.62 (+/- 0.02)&-86.3&< \textbf{0.0001}\\
Java&-1.29 (+/- 0.02)&-72.5&< \textbf{0.0001}\\
Julia&-4.90 (+/- 0.04)&-132.6&< \textbf{0.0001}\\
JavaScript&-0.97 (+/- 0.02)&-56.1&< \textbf{0.0001}\\
Lua&-2.21 (+/- 0.02)&-111.2&< \textbf{0.0001}\\
Perl&-2.43 (+/- 0.02)&-118.2&< \textbf{0.0001}\\
PHP&-1.76 (+/- 0.02)&-93.6&< \textbf{0.0001}\\
R&-2.80 (+/- 0.02)&-128.2&< \textbf{0.0001}\\
Ruby&-1.87 (+/- 0.02)&-98.4&< \textbf{0.0001}\\
Racket&-3.41 (+/- 0.02)&-138.2&< \textbf{0.0001}\\
Rust&-2.74 (+/- 0.02)&-125.3&< \textbf{0.0001}\\
Scala&-1.92 (+/- 0.02)&-100.3&< \textbf{0.0001}\\
Swift&-2.05 (+/- 0.02)&-105.3&< \textbf{0.0001}\\
TypeScript&-1.11 (+/- 0.02)&-63.6&< \textbf{0.0001}\\\hline
    \end{tabular}
    \caption{Mixed-effects results for InCoder \newHE language comparison}\label{tab:mem_he_lang_incoder}
\end{table}

\subsection{\newMBPP Mixed-Effects Results from \Cref{subsec:newmbpp-results}} 

To quantify the differences in performance among programming languages, a model with a fixed effect of programming language and random effects for problem number was fitted to the Codex pass@1 \newMBPP data. Dummy coding was used with Python as the reference level; slopes for each language indicate differences between the pass@1 rate for Python and that language.

Table \ref{tab:mbpp_mem_lang_codex} shows the full estimates found by the model.

\begin{table}[H]
    \centering
    \begin{tabular}{|llll|}\hline
Fixed effects&$\widehat{\beta}$&$z$&$p$\\\hline
(Intercept)&0.28 (+/- 0.2)&1.74&0.08\\  
Bash&-2.9 (+/- 0.02)&-193.22&< \textbf{0.0001}\\
C++&-1.16 (+/- 0.01)&-82.73&< \textbf{0.0001}\\
C\#&-1.80 (+/- 0.01)&-126.81&< \textbf{0.0001}\\
D&-2.47 (+/- 0.01)&-169.64&< \textbf{0.0001}\\
Go&-1.22 (+/- 0.01)&-86.33&< \textbf{0.0001}\\
Java&-1.14 (+/- 0.01)&-81.19&< \textbf{0.0001}\\
Julia&-1.72 (+/- 0.01)&-121.68&< \textbf{0.0001}\\
JavaScript&-0.36 (+/- 0.01)&-24.96&< \textbf{0.0001}\\
Lua&-0.94 (+/- 0.01)&-66.54&< \textbf{0.0001}\\
Perl&-3.06 (+/- 0.02)&-201.65&< \textbf{0.0001}\\
PHP&-1.03 (+/- 0.01)&-72.81&< \textbf{0.0001}\\
R&-2.48 (+/- 0.01)&-169.93&< \textbf{0.0001}\\
Ruby&-0.85 (+/- 0.01)&-60.10&< \textbf{0.0001}\\
Racket&-2.39 (+/- 0.01)&-164.56&< \textbf{0.0001}\\
Rust&-1.69 (+/- 0.01)&-119.60&< \textbf{0.0001}\\
Scala&-1.20 (+/- 0.01)&-85.31&< \textbf{0.0001}\\
Swift&-1.66 (+/- 0.01)&-117.36&< \textbf{0.0001}\\
TypeScript&-0.87 (+/- 0.01)&-61.89&< \textbf{0.0001}\\\hline
    \end{tabular}
    \caption{Mixed-effects results for Codex \newMBPP language comparison}\label{tab:mbpp_mem_lang_codex}
\end{table}

A similar model was fit to the CodeGen pass@1 \newMBPP data. Table \ref{tab:mbpp_mem_lang_codegen} shows the full estimates found by the model.

\begin{table}[H]
    \centering
    \begin{tabular}{|llll|}\hline
Fixed effects&$\widehat{\beta}$&$z$&$p$\\\hline
(Intercept)&-2.85 (+/- 0.2)&-13.9&< \textbf{0.0001}\\
Bash&-5.74 (+/- 0.04)&-129.1&< \textbf{0.0001}\\
C++&-0.31 (+/- 0.02)&-19.6&< \textbf{0.0001}\\
C\#&-1.68 (+/- 0.02)&-97.7&< \textbf{0.0001}\\
D&-1.63 (+/- 0.02)&-95.3&< \textbf{0.0001}\\
Go&-0.97 (+/- 0.02)&-59.8&< \textbf{0.0001}\\
Java&-0.22 (+/- 0.02)&-13.6&< \textbf{0.0001}\\
Julia&-9.23 (+/- 0.2)&-43.0&< \textbf{0.0001}\\
JavaScript&0.16 (+/- 0.02)&9.9&< \textbf{0.0001}\\
Lua&-2.60 (+/- 0.02)&-134.2&< \textbf{0.0001}\\
Perl&-2.88 (+/- 0.02)&-141.8&< \textbf{0.0001}\\
PHP&-1.70 (+/- 0.02)&-98.7&< \textbf{0.0001}\\
R&-2.34 (+/- 0.02)&-125.7&< \textbf{0.0001}\\
Ruby&-22.34 (+/- 149.3)&-0.2&0.881\\    
Racket&-5.57 (+/- 0.04)&-133.2&< \textbf{0.0001}\\
Rust&-4.03 (+/- 0.03)&-154.5&< \textbf{0.0001}\\
Scala&-3.59 (+/- 0.02)&-153.1&< \textbf{0.0001}\\
Swift&-3.97 (+/- 0.03)&-154.5&< \textbf{0.0001}\\
TypeScript&-0.08 (+/- 0.02)&-4.8&< \textbf{0.0001}\\\hline
    \end{tabular}
    \caption{Mixed-effects results for CodeGen \newMBPP language comparison}\label{tab:mbpp_mem_lang_codegen}
\end{table}

A similar model was fit to the InCoder pass@1 data. Table \ref{tab:mbpp_mem_lang_incoder} shows the full estimates found by the model.

\begin{table}[H]
    \centering
    \begin{tabular}{|llll|}\hline
Fixed effects&$\widehat{\beta}$&$z$&$p$\\\hline
(Intercept)&-3.85 (+/- 0.2)&-22.67&< \textbf{0.0001}\\
Bash&-2.09 (+/- 0.02)&-87.5&< \textbf{0.0001}\\
C++&-0.18 (+/- 0.02)&-10.1&< \textbf{0.0001}\\
C\#&-0.88 (+/- 0.02)&-46.0&< \textbf{0.0001}\\
D&-1.60 (+/- 0.02)&-74.5&< \textbf{0.0001}\\
Go&-0.34 (+/- 0.02)&-19.2&< \textbf{0.0001}\\
Java&-0.07 (+/- 0.02)&-3.8&\textbf{0.0001}\\
Julia&-1.55 (+/- 0.02)&-73.0&< \textbf{0.0001}\\
JavaScript&1.65 (+/- 0.04)&39.7&< \textbf{0.0001}\\
Lua&-0.36 (+/- 0.02)&-20.0&< \textbf{0.0001}\\
Perl&-0.60 (+/- 0.02)&-32.4&< \textbf{0.0001}\\
PHP&0.54 (+/- 0.02)&32.0&< \textbf{0.0001}\\
R&-0.89 (+/- 0.02)&-46.7&< \textbf{0.0001}\\
Ruby&-0.29 (+/- 0.02)&-16.6&< \textbf{0.0001}\\
Racket&-1.88 (+/- 0.02)&-82.6&< \textbf{0.0001}\\
Rust&-1.157 (+/- 0.02)&-58.3&< \textbf{0.0001}\\
Scala&-1.11 (+/- 0.02)&-56.4&< \textbf{0.0001}\\
Swift&-0.60 (+/- 0.02)&-32.6&< \textbf{0.0001}\\
TypeScript&0.84968 (+/- 0.02)&51.0&< \textbf{0.0001}\\\hline
    \end{tabular}
    \caption{Mixed-effects results for InCoder \newMBPP language comparison}\label{tab:mbpp_mem_lang_incoder}
\end{table}

\subsection{Mixed-Effects Results for \Cref{subsubsec:he_freq} and \Cref{subsubsec:mbpp_freq}}

A mixed-effects model treating Frequency as a fixed-effects was fit to the Codex \newHE data, with random effects for language and problem. Table \ref{tab:he_mem_freq} shows the full estimates found by the model.

\begin{table}[H]
    \centering
    \begin{tabular}{|llll|}\hline
Fixed effects&$\widehat{\beta}$&$z$&$p$\\\hline
(Intercept)&-0.31 (+/- 0.2)&-1.3&0.19\\
Niche&-1.68 (+/-0.4)&-4.5&< \textbf{0.0001}\\ 
Low&-1.73 (+/- 0.3)&-5.24&< \textbf{0.0001}\\
Medium&-0.85 (+/- 0.3)&-2.8&\textbf{0.006}\\\hline
    \end{tabular}
    \caption{Mixed-effects results for \newHE Codex language frequency comparison}\label{tab:he_mem_freq}
\end{table}

A mixed-effects model treating Frequency as a fixed-effects was fit to the Codex \newMBPP data, with random effects for language and problem. Table \ref{tab:mbbp_mem_freq} shows the full estimates found by the model.

\begin{table}[H]
    \centering
    \begin{tabular}{|llll|}\hline
Fixed effects&$\widehat{\beta}$&$z$&$p$\\\hline
(Intercept)&-0.19 (+/- 0.3)&-0.6&0.56\\    
Low&-2.13 (+/- 0.4)&-4.8&< \textbf{0.0001}\\
Medium&-0.93 (+/- 0.3)&-2.72&\textbf{0.007}\\ 
Niche&-1.74 (+/- 0.4)&-4.10&< \textbf{0.0001}\\\hline
    \end{tabular}
    \caption{Mixed-effects results for \newMBPP Codex language frequency comparison}\label{tab:mbbp_mem_freq}
\end{table}

\subsection{Mixed-Effects Results for \Cref{subsec:ablation}} 

Three mixed-effects models were fit to the \newHE data for the ablation study.

A mixed-effects model was fit to the InCoder pass@1 rates to explore how the translation components affect its performance.  This model compared InCoder pass@1 rates for four experiments: Doctest-Only Translation, Full Translation, No Translation, and Remove Doctests. Experiment was treated as a fixed-effect, with Python and Doctest-Only Translation as the reference levels. Random effects for language were included; random effects for problem were not included, as the extremely low pass rates for many problems caused instability in estimating them. Table \ref{tab:mem_ablation_small_incoder} shows the full estimates found by the model.

\begin{table}[H]
    \centering
    \begin{tabular}{|llll|}\hline
Fixed effects&$\widehat{\beta}$&$z$&$p$\\\hline
(Intercept)&-2.97 (+/- 0.2)&-15.6&< \textbf{0.001}\\
Remove&0.20 (+/- 0.07)&2.8&\textbf{0.005}\\ 
No Translation&0.03 (+/- 0.03)&1.0&0.32\\    
Full Translation&0.02 (+/- 0.02)&1.3&0.20\\\hline  
    \end{tabular}
    \caption{Mixed-effects results for the InCoder ablation study}\label{tab:mem_ablation_small_incoder}
\end{table}

A similar mixed-effects model was fit to understand the impact of translating natural language terms and doctests on Codex performance. This model compared Codex pass@1 rates for four experiments: Doctest-Only Translation, Full Translation, No Translation, and Remove Doctests. Experiment was treated as a fixed-effect, with Python and Doctest-Only Translation as the reference levels. Random effects for problem and language were included. Table \ref{tab:mem_ablation_small} shows the full estimates found by the model.

\begin{table}[H]
    \centering
    \begin{tabular}{|llll|}\hline
Fixed effects&$\widehat{\beta}$&$z$&$p$\\\hline
(Intercept)&-1.24 (+/- 0.3)&-4.4&< \textbf{0.0001}\\
Full Translation&0.04 (+/- 0.02)&2.2&\textbf{0.03}\\  
No Translation&-0.08 (+/- 0.1)&-1.3&0.2\\    
Remove&-0.35 (+/- 0.1)&-3.8&< \textbf{0.0001}\\\hline
    \end{tabular}
    \caption{Mixed-effects results for the Codex ablation study}\label{tab:mem_ablation_small}
\end{table}

A second model was fitted for Codex treating both Language and Experiment as fixed-effects, with interaction terms included. For this model, we include only random intercepts but not random slopes for Problem, because of the large number of effects the model must estimate. Tables \ref{tab:mem_ablation} and \ref{tab:mem_ablation_interaction} show the full estimates found by the model.

\begin{table}[H]
    \centering
    \begin{tabular}{|llll|}\hline
Fixed effects&$\widehat{\beta}$&$z$&$p$\\\hline\hline
(Intercept)&-0.44 (+/- 0.2)&-2.2&\textbf{0.03}\\  
Full Translation&-0.006 (+/- 0.02)&-0.3&0.78\\
No Translation&-0.07 (+/- 0.02)&-3.1&\textbf{0.002}\\ 
Remove&-0.14 (+/- 0.02)&-6.6&< \textbf{0.0001}\\\hline\hline
Bash&-1.75 (+/- 0.02)&-77.8&< \textbf{0.0001}\\
C++&0.17 (+/- 0.02)&8.1&< \textbf{0.0001}\\
C\#&-1.45 (+/- 0.02)&-65.7&< \textbf{0.0001}\\
D&-1.97 (+/- 0.02)&-86.5&< \textbf{0.0001}\\
Go&-1.14 (+/- 0.02)&-52.5&< \textbf{0.0001}\\
Java&-0.63 (+/- 0.02)&-29.2&< \textbf{0.0001}\\
Julia&-0.87 (+/- 0.02)&-40.4&< \textbf{0.0001}\\
JavaScript&0.15 (+/- 0.02)&7.2&< \textbf{0.0001}\\
Lua&-0.48 (+/- 0.02)&-22.7&< \textbf{0.0001}\\
Perl&-1.10 (+/- 0.02)&-51.0&< \textbf{0.0001}\\
PHP&-0.006 (+/- 0.02)&-0.3&0.76\\    
R&-2.02 (+/- 0.02)&-88.8&< \textbf{0.0001}\\
Ruby&-0.31 (+/- 0.02)&-14.4&< \textbf{0.0001}\\
Racket&-2.36 (+/- 0.02)&-100.3&< \textbf{0.0001}\\
Rust&-0.30 (+/- 0.02)&-14.0&< \textbf{0.0001}\\
Scala&-0.24 (+/- 0.02)&-11.0&< \textbf{0.0001}\\
Swift&-0.62 (+/- 0.02)&-28.9&< \textbf{0.0001}\\
TypeScript&0.10 (+/- 0.02)&4.8&< \textbf{0.0001}\\\hline
\end{tabular}
    \caption{Mixed-effects results for the Codex ablation study by language, main effects}\label{tab:mem_ablation}
\end{table}

\begin{table}[H]
    \centering
 \begin{tabular}{|llll|}\hline
Fixed effects&$\widehat{\beta}$&$z$&$p$\\\hline\hline
Full Translation*Bash&-0.02 (+/- 0.03)&-0.5&0.58\\    
No Translation*Bash&-0.47 (+/- 0.03)&-14.4&< \textbf{0.0001}\\
Remove*Bash&-0.59 (+/- 0.03)&-17.9&< \textbf{0.0001}\\\hline
Full Translation*C++&-0.03 (+/- 0.03)&-0.9&0.35\\    
No Translation*C++&-0.15 (+/- 0.03)&-5.0&< \textbf{0.0001}\\
Remove*C++&-0.11 (+/- 0.03)&-3.7&\textbf{0.0002}\\\hline
Full Translation*C\#&-0.02 (+/- 0.03)&-0.6&0.58\\    
No Translation*C\#&0.05 (+/- 0.03)&1.6&0.10\\    
Remove*C\#&0.2 (+/- 0.03)&6.9&< \textbf{0.0001}\\\hline
Full Translation*D&0.02 (+/- 0.03)&0.5&0.59\\    
No Translation*D&0.20 (+/- 0.03)&6.4&< \textbf{0.0001}\\
Remove*D&0.10 (+/- 0.03)&3.2&\textbf{0.001}\\ \hline
Full Translation*Go&-0.03 (+/- 0.03)&-0.8&0.41\\    
No Translation*Go&-0.03 (+/- 0.03)&-1.0&0.32\\    
Remove*Go&0.05 (+/- 0.03)&1.8&0.08\\  \hline
Full Translation*Java&0.12 (+/- 0.03)&3.9&< \textbf{0.0001}\\
No Translation*Java&0.14 (+/- 0.03)&4.5&< \textbf{0.0001}\\
Remove*Java&0.12 (+/- 0.03)&4.0&< \textbf{0.0001}\\\hline
Full Translation*Julia&0.05 (+/- 0.03)&1.8&0.07\\  
No Translation*Julia&0.05 (+/- 0.03)&1.7&0.10\\  
Remove*Julia&-0.09 (+/- 0.03)&-2.9&\textbf{0.004}\\ \hline
Full Translation*JavaScript&0.01 (+/- 0.03)&0.4&0.66\\    
No Translation*JavaScript&0.08 (+/- 0.03)&2.6&\textbf{0.01}\\  
Remove*JavaScript&-0.09 (+/-0.03)&-2.8&\textbf{0.005}\\\hline 
Full Translation*Lua&0.06 (+/- 0.03)&1.9&0.06\\  
No Translation*Lua&-0.04 (+/- 0.03)&-1.3&0.19\\    
Remove*Lua&-0.12 (+/- 0.03)&-3.8&\textbf{0.0001}\\\hline
Full Translation*Perl&0.23 (+/- 0.03)&7.5&< \textbf{0.0001}\\
No Translation*Perl&-0.25 (+/- 0.03)&-8.1&< \textbf{0.0001}\\
Remove*Perl&-0.21 (+/- 0.03)&-6.9&< \textbf{0.0001}\\\hline
Full Translation*PHP&0.06 (+/- 0.03)&1.8&0.07\\  
No Translation*PHP&0.02 (+/- 0.03)&0.6&0.58\\    
Remove*PHP&-0.26 (+/- 0.03)&-8.6&< \textbf{0.0001}\\\hline
Full Translation*R&0.22 (+/- 0.03)&7.0&< \textbf{0.0001}\\
No Translation*R&0.26 (+/- 0.03)&8.1&< \textbf{0.0001}\\
Remove*R&-0.11 (+/- 0.03)&-3.5&{0.0004}\\\hline
Full Translation*Ruby&0.012 (+/- 0.03)&0.4&0.68\\    
No Translation*Ruby&0.02 (+/- 0.03)&0.5&0.58\\    
Remove*Ruby&-0.19 (+/- 0.03)&-6.1&< \textbf{0.0001}\\\hline
Full Translation*Racket&0.04 (+/- 0.03)&1.2&0.23\\    
No Translation*Racket&-0.07 (+/- 0.03)&-2.0&0.05\\  
Remove*Racket&-0.21 (+/- 0.03)&-6.1&< \textbf{0.0001}\\\hline
Full Translation*Rust&0.03 (+/- 0.03)&1.0&0.31\\    
No Translation*Rust&-0.04 (+/- 0.03)&-1.5&0.14\\    
Remove*Rust&-0.34 (+/- 0.03)&-11.2&< \textbf{0.0001}\\\hline
Full Translation*Scala&0.01 (+/- 0.03)&0.5&0.64\\    
No Translation*Scala&0.17 (+/- 0.03)&5.5&< \textbf{0.0001}\\
Remove*Scala&0.03 (+/- 0.03)&1.1&0.26\\\hline    
Full Translation*Swift&-0.01 (+/- 0.03)&-0.5&0.63\\    
No Translation*Swift&-0.39 (+/- 0.03)&-13.0&< \textbf{0.0001}\\
Remove*Swift&-0.24 (+/- 0.03)&-8.0&< \textbf{0.0001}\\\hline
Full Translation*TypeScript&0.05 (+/- 0.03)&1.6&0.11\\    
No Translation*TypeScript&0.11 (+/- 0.03)&3.8&\textbf{0.0002}\\
Remove*TypeScript&-0.25 (+/- 0.03)&-8.4&< \textbf{0.0001}\\\hline\hline
\end{tabular}
    \caption{Mixed-effects results for the Codex ablation study by language, interaction effects}\label{tab:mem_ablation_interaction}
\end{table}

\subsection{Mixed-Effects Results from \Cref{subsec:static-types}} 

A mixed-effects model treating Static Type-checking as a fixed-effects was fit to the \newHE data, with random effects for language and problem. Interaction terms were included for Typed with each frequency category. Table \ref{tab:he_mem_types} shows the full estimates found by the model.

\begin{table}[H]
    \centering
    \begin{tabular}{|llll|}\hline
Fixed effects&$\widehat{\beta}$&$z$&$p$\\\hline
(Intercept)&-1.31 (+/- 0.3)&-3.94&< \textbf{0.0001}\\
Typed&-0.36 (+/- 0.4)&-0.95&0.34\\\hline 
    \end{tabular}
    \caption{Mixed-effects results for \newHE Codex static type-checking comparison}\label{tab:he_mem_types}
\end{table}

A similar mixed-effects model was fit to the Codex \newMBPP data. Table \ref{tab:mbpp_mem_types} shows the full estimates found by the model.

\begin{table}[H]
    \centering
    \begin{tabular}{|llll|}\hline
Fixed effects&$\widehat{\beta}$&$z$&$p$\\\hline
(Intercept)&-1.31 (+/- 0.4)&-3.3&\textbf{0.0009}\\
Typed&-0.52 (+/- 0.4)&-1.2&0.23\\\hline
    \end{tabular}
    \caption{Mixed-effects results for \newMBPP Codex static type-checking comparison}\label{tab:mbpp_mem_types}
\end{table}

A mixed-effects model testing the effect of removing Python type annotations was fit to the Codex \newHE data. Annotations was treated as a fixed-effect and problem as a random effect. Table \ref{tab:mem_python_types} shows the full estimates found by the model.

\begin{figure}[H]
    \centering
    \includegraphics[width=0.5\textwidth]{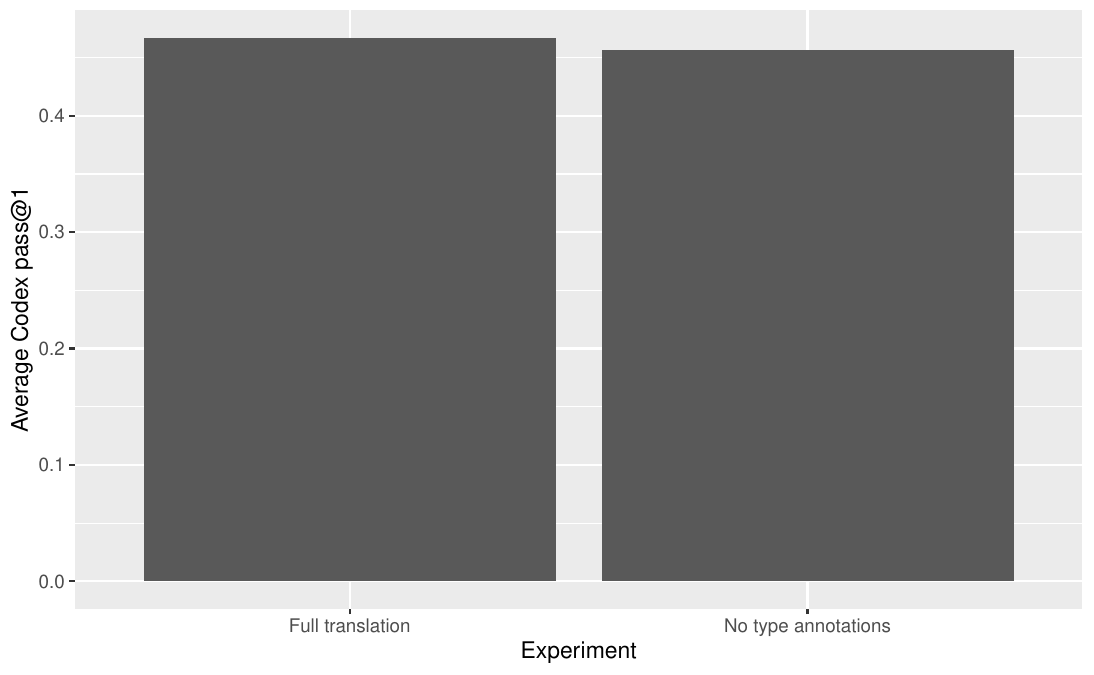}
    \caption{Impact of Python type annotations on Codex performance}
    \label{fig:pyexpt}
\end{figure}

\begin{table}[H]
    \centering
    \begin{tabular}{|llll|}\hline
Fixed effects&$\widehat{\beta}$&$z$&$p$\\\hline
Intercept&-0.26 (+/- 0.5)&-0.5&0.60\\
Annotations&-0.21 (+/- ).2)&-1.2&0.22\\\hline
    \end{tabular}
    \caption{Mixed-effects results for Python type annotation experiments}\label{tab:mem_python_types}
\end{table}

A mixed-effects model testing the effect of weakening TypeScript annotations to Any and running without static type-checking was fit. There were three fixed-effects: Any, comparing TypeScript with precise types to TypeScript with all Any types; JS, comparing TypeScript with annotations to JavaScript; and NoCheck, comparing TypeScript with and without static type-checking. 
Table \ref{tab:mem_script} shows the full estimates found by the model.

\begin{figure}[H]
    \centering
    \includegraphics[width=0.75\textwidth]{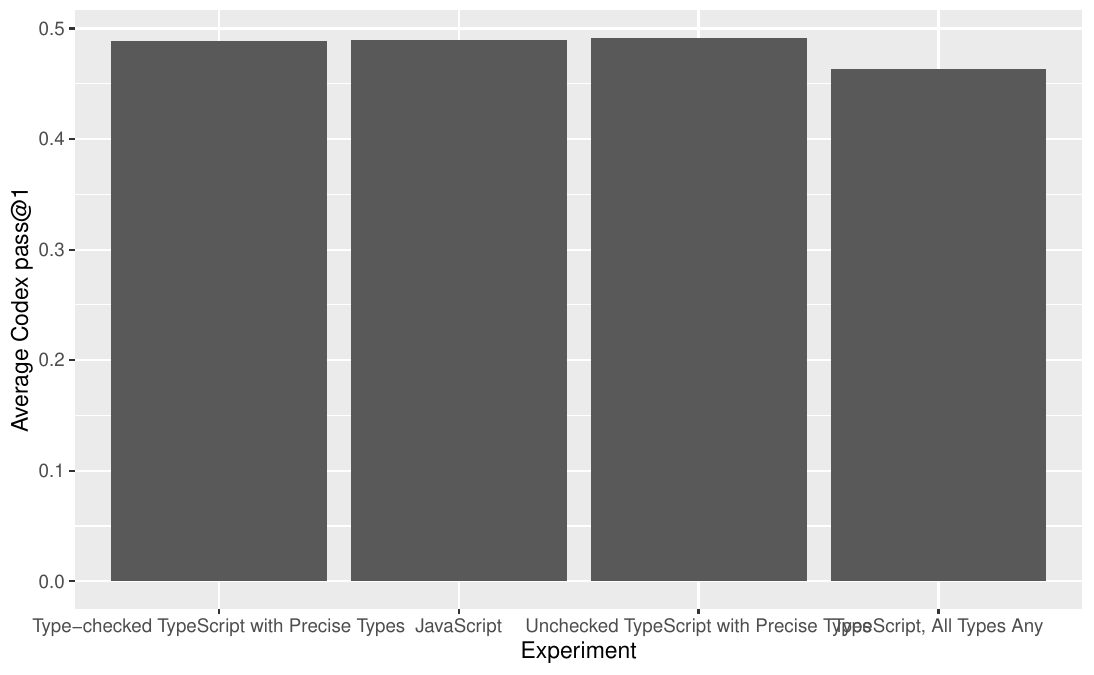}
    \caption{Impact of type-checking and precise type annotations on TypeScript performance}
    \label{fig:tsexpt}
\end{figure}

\begin{table}[H]
    \centering
    \begin{tabular}{|llll|}\hline
Fixed effects&$\widehat{\beta}$&$z$&$p$\\\hline
Intercept&-0.24 (+/- 0.4)&-0.6&0.56\\
JavaScript&-0.03 (+/- 0.03)&-1.2&0.23\\
Any Types&-0.38 (+/- 0.03)&-13.3&< \textbf{0.001}\\
NoCheck&0.04 (+/- 0.03)&1.5&0.14\\\hline
    \end{tabular}
    \caption{Mixed-effects results for TypeScript experiments}\label{tab:mem_script}
\end{table}

\subsection{Mixed-Effects Results from \Cref{subsec:prompt-engineering}} 

Tables \ref{tab:mem_comments_php} and \ref{tab:mem_comments_rkt} shows the results of singe-line versus multi-line comments for PHP and Racket. Separate models were run for each language, with multi-line as a fixed effect and problem number as a random effect.

\begin{figure}[H]
    \centering
    \includegraphics[width=0.5\textwidth]{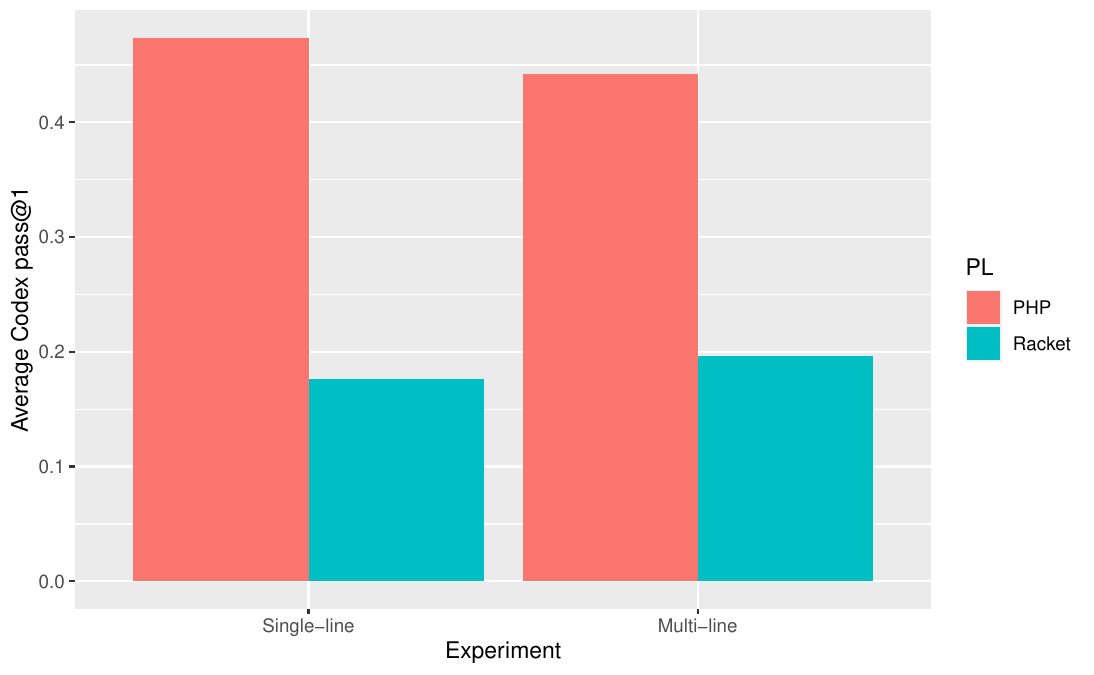}
    \caption{Impact of comment style on Codex performance for PHP and Racket}
    \label{fig:multiexpt}
\end{figure}

\begin{table}[H]
    \centering
    \begin{tabular}{|llll|}\hline
Fixed effects&$\widehat{\beta}$&$z$&$p$\\\hline
Intercept&-0.46 (+/- 0.4)&-1.2&0.22\\
Multi-line&-0.43 (+/-0.1)&-3.3&\textbf{0.001}\\\hline
    \end{tabular}
    \caption{Mixed-effect model estimates for PHP comment experiment}\label{tab:mem_comments_php}
\end{table}

\begin{table}[H]
    \centering
    \begin{tabular}{|llll|}\hline
Fixed effects&$\widehat{\beta}$&$z$&$p$\\\hline
Intercept&-4.62 (+/- 0.4)&-10.9&< \textbf{0.0001}\\
Multi-line&1.26 (+/-0.2)&6.4&< \textbf{0.0001}\\\hline
    \end{tabular}
    \caption{Mixed-effect model estimates for Racket comment experiment}\label{tab:mem_comments_rkt}
\end{table}

Table \ref{tab:mem_perl} shows the results of comparing Perl with and without an argument-naming line after the function signature. Argument-naming was treated as a fixed effect and problem number as a random effect.

\begin{figure}[H]
    \centering
    \includegraphics[width=0.5\textwidth]{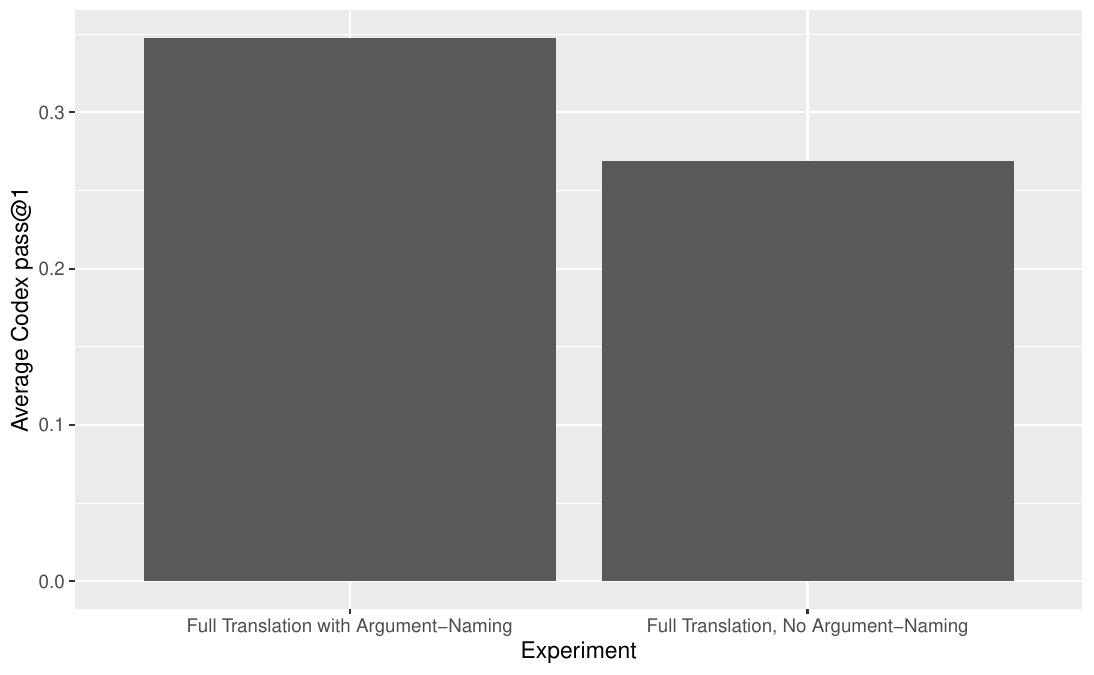}
    \caption{Impact of argument-naming line on Codex performance for Perl}
    \label{fig:perlexpt}
\end{figure}

\begin{table}[H]
    \centering
    \begin{tabular}{|llll|}\hline
Fixed effects&$\widehat{\beta}$&$z$&$p$\\\hline
Intercept&-3.03 (+/- 0.4)&-7.9&<\textbf{0.0001}\\
Argument-naming&0.81 (+/-0.2)&3.6&\textbf{0.0008}\\\hline
    \end{tabular}
    \caption{Mixed-effect model estimates for Perl experiment}\label{tab:mem_perl}
\end{table}

Table \ref{tab:mem_bash} shows the results of comparing Bash with and without encoding-specifying comments. Comments and NL Translation were treated as fixed effects and problem number as a random effect; an interaction term for Comments and NL Translation was also included.

\begin{figure}[H]
    \centering
    \includegraphics[width=0.5\textwidth]{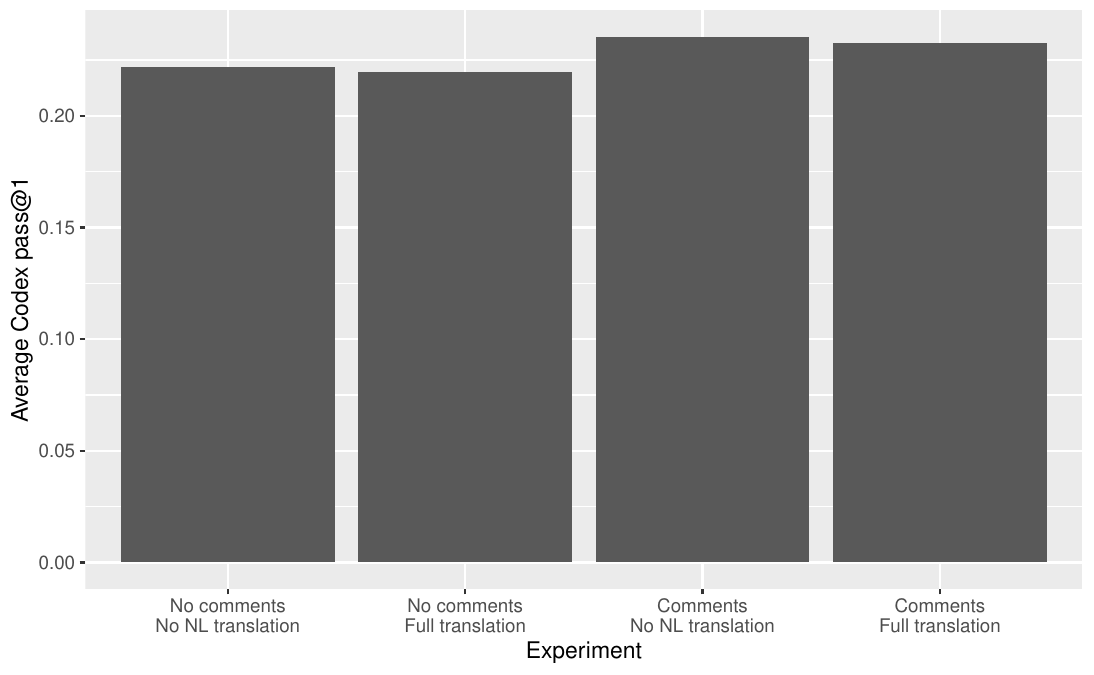}
    \caption{Impact of encoding comments and NL translation on Codex performance for Bash}
    \label{fig:bashexpt}
\end{figure}

\begin{table}[H]
    \centering
    \begin{tabular}{|llll|}\hline
Fixed effects&$\widehat{\beta}$&$z$&$p$\\\hline
Intercept&-3.09 (+/- 0.3)&-9.9&< \textbf{0.001}\\
Comments&0.01 (+/- 0.1)&0.08&0.94\\
Rewording&-0.04 (+/- 0.03)&-1.3&0.19\\
Comments*Rewording&0.08 (+/-0.4)&1.8&0.07\\\hline
    \end{tabular}
    \caption{Mixed-effect model estimates for Bash experiment}\label{tab:mem_bash}
\end{table}

\subsection{Mixed-Effects Results for \Cref{subsec:features}} 

We categorize problems into groups based on which Python language features they use: dictionaries, tuples, Booleans, lists, or none of the above. We base these categorizations on the Python type annotations for each problem. Problems were coded 1 Tuple, List, Bool, and Dictionary if they contain a type annotation for the respective feature, and 0 otherwise. 

We fit a mixed-effects model to understand how Codex pass@1 rates are affected by the language features used in the problem, using Tuple, List, Bool, and Dictionary as fixed-effects, with random effects for problem and language. Table \ref{tab:mem_features_small} shows the full estimates found by the model.

\begin{table}[H]
    \centering
    \begin{tabular}{|llll|}\hline
Fixed effects&$\widehat{\beta}$&$z$&$p$\\\hline
(Intercept)&-1.19 (+/- 0.3)&-3.5&< \textbf{0.001}\\
List&-0.15 (+/- 0.4)&-0.3&0.73\\    
Bool&0.10 (+/- 0.6)&0.2&0.86\\    
Tuple&-0.73 (+/- 0.9)&-0.8&0.40\\    
Dictionary&-3.27 (+/- 1.8)&-1.8&0.07\\\hline
    \end{tabular}
    \caption{Mixed-effects results for the impact of language features}\label{tab:mem_features_small}
\end{table}

A second model was fit with interaction effects for languages and language features. Table \ref{tab:mem_features_large}-\ref{tab:mem_features_large2} shows the full estimates found by the model.

\begin{table}[H]
    \centering
    \begin{tabular}{|llll|}\hline
Fixed effects&$\widehat{\beta}$&$z$&$p$\\\hline  
(Intercept)&-0.37 (+/- 0.3)&-1.1&0.26\\   
Bash&-1.61 (+/- 0.2)&-86.6&<\textbf{0.0001}\\
C++&-0.45 (+/- 0.2)&-25.4&<\textbf{0.0001}\\
C\#&-0.86 (+/- 0.2)&-47.7&<\textbf{0.0001}\\
D&-2.03 (+/- 0.2)&-106.6&<\textbf{0.0001}\\
Go&-0.96 (+/- 0.2)&-53.0&<\textbf{0.0001}\\
Java&-0.42 (+/- 0.2)&-23.8&<\textbf{0.0001}\\
Julia&-1.25 (+/- 0.2)&-68.5&<\textbf{0.0001}\\
JavaScript&-0.07 (+/- 0.2)&-3.8&\textbf{0.0001}\\
Lua&-0.32 (+/- 0.2)&-18.4&<\textbf{0.0001}\\
Perl&-0.54 (+/- 0.2)&-30.6&<\textbf{0.0001}\\
PHP&-0.37 (+/- 0.2)&-20.6&<\textbf{0.0001}\\
R&-1.79 (+/- 0.2)&-96.1&<\textbf{0.0001}\\
Ruby&-0.63 (+/- 0.2)&-35.6&<\textbf{0.0001}\\
Racket&-2.24 (+/- 0.2)&-114.7&<\textbf{0.0001}\\
Rust&-0.84 (+/- 0.2)&-46.7&<\textbf{0.0001}\\
Scala&-0.49 (+/- 0.2)&-27.2&<\textbf{0.0001}\\
Swift&-1.60 (+/- 0.2)&-86.5&<\textbf{0.0001}\\
TypeScript&-0.15 (+/- 0.2)&-8.4&<\textbf{0.0001}\\
List&-0.20 (+/- 0.4)&-0.5&0.62\\    
Bool&0.07 (+/- 0.5)&0.1&0.9\\    
Tuple&-0.39 (+/- 0.9)&-0.44&0.66\\    
Dictionary&-1.65 (+/- 1.9)&-0.89&0.37\\    
Bash:List&-0.48 (+/- 0.02)&-20.2&<\textbf{0.0001}\\
C++:List&0.78 (+/- 0.02)&34.6&<\textbf{0.0001}\\
C\#:List&-1.17 (+/- 0.02)&-50.9&<\textbf{0.0001}\\
D:List&0.055 (+/- 0.02)&2.3&\textbf{0.019}\\  
Go:List&-0.20 (+/- 0.02)&-8.8&<\textbf{0.0001}\\
Java:List&0.15 (+/- 0.02)&6.8&<\textbf{0.0001}\\
Julia:List&0.40 (+/- 0.02)&17.6&<\textbf{0.0001}\\
JavaScript:List&0.28 (+/- 0.02)&12.4&<\textbf{0.0001}\\
Lua:List&-0.42 (+/- 0.02)&-19.2&<\textbf{0.0001}\\
Perl:List&-0.72 (+/- 0.02)&-32.2&<\textbf{0.0001}\\
PHP:List&0.27 (+/- 0.02)&12.2&<\textbf{0.0001}\\
R:List&-0.47 (+/- 0.02)&-19.9&<\textbf{0.0001}\\
Ruby:List&0.49 (+/- 0.02)&22.2&<\textbf{0.0001}\\
Racket:List&-0.01 (+/- 0.02)&-0.48&0.63\\    
Rust:List&0.62 (+/- 0.02)&27.5&<\textbf{0.0001}\\
Scala:List&0.48 (+/- 0.02)&21.5&<\textbf{0.0001}\\
Swift:List&1.30 (+/- 0.02)&56.9&<\textbf{0.0001}\\
TypeScript:List&0.34 (+/- 0.02)&15.4&<\textbf{0.0001}\\
Bash:Bool&-0.52 (+/- 0.03)&-17.04&<\textbf{0.0001}\\
C++:Bool&0.87 (+/- 0.03)&28.8&<\textbf{0.0001}\\
C\#:Bool&0.52 (+/- 0.02)&18.2&<\textbf{0.0001}\\
D:Bool&0.44 (+/- 0.03)&15.2&<\textbf{0.0001}\\
Go:Bool&-0.16 (+/- 0.03)&-5.7&<\textbf{0.0001}\\
Java:Bool&-0.49 (+/- 0.03)&-17.0&<\textbf{0.0001}\\
Julia:Bool&0.81 (+/- 0.03)&28.2&<\textbf{0.0001}\\
JavaScript:Bool&0.25 (+/- 0.03)&8.5&<\textbf{0.0001}\\
Lua:Bool&0.11 (+/- 0.03)&3.9&<\textbf{0.0001}\\
Perl:Bool&-1.18 (+/- 0.03)&-40.2&<\textbf{0.0001}\\
PHP:Bool&0.79 (+/- 0.03)&27.2&<\textbf{0.0001}\\
R:Bool&0.42 (+/- 0.03)&14.6&<\textbf{0.0001}\\
Ruby:Bool&0.074 (+/- 0.03)&2.6&\textbf{0.009}\\ 
Racket:Bool&-1.10 (+/- 0.03)&-32.2&<\textbf{0.0001}\\
Rust:Bool&0.60 (+/- 0.03)&20.9&<\textbf{0.0001}\\
Scala:Bool&0.18 (+/- 0.03)&6.2&<\textbf{0.0001}\\
Swift:Bool&0.47 (+/- 0.03)&15.9&<\textbf{0.0001}\\
TypeScript:Bool&0.14 (+/- 0.03)&5.0&<\textbf{0.0001}\\\hline
\end{tabular}
\caption{Mixed-effects results for the impact of language features}\label{tab:mem_features_large}
\end{table}

\begin{table}[H]
    \centering
    \begin{tabular}{|llll|}\hline
Fixed effects&$\widehat{\beta}$&$z$&$p$\\\hline  

Bash:Tuple&-2.17 (+/- 0.09)&-24.9&<\textbf{0.0001}\\
C++:Tuple&0.35 (+/- 0.05)&7.8&<\textbf{0.0001}\\
C\#:Tuple&-0.71 (+/- 0.05)&-13.6&<\textbf{0.0001}\\
D:Tuple&-0.043 (+/- 0.05)&-0.8&0.43\\    
Go:Tuple&-1.22 (+/- 0.06)&-22.1&<\textbf{0.0001}\\
Java:Tuple&-2.44 (+/- 0.06)&-39.9&<\textbf{0.0001}\\
Julia:Tuple&0.24 (+/- 0.05)&4.9&<\textbf{0.0001}\\
JavaScript:Tuple&0.80 (+/- 0.04)&17.8&<\textbf{0.0001}\\
Lua:Tuple&0.078 (+/- 0.04)&1.7&0.08\\  
Perl:Tuple&-0.21 (+/- 0.05)&-4.4&<\textbf{0.0001}\\
PHP:Tuple&0.50 (+/- 0.04)&11.3&<\textbf{0.0001}\\
R:Tuple&0.07 (+/- 0.05)&1.4&0.16\\    
Ruby:Tuple&0.21 (+/- 0.04)&4.7&<\textbf{0.0001}\\
Racket:Tuple&0.22 (+/- 0.06)&4.0&<\textbf{0.0001}\\
Rust:Tuple&-0.20 (+/- 0.05)&-4.2&<\textbf{0.0001}\\
Scala:Tuple&0.41 (+/- 0.04)&9.2&<\textbf{0.0001}\\
Swift:Tuple&0.32 (+/- 0.05)&6.7&<\textbf{0.0001}\\
TypeScript:Tuple&0.73 (+/- 0.05)&15.4&<\textbf{0.0001}\\
Bash:Dictionary&-13.29 (+/- 115.6)&-0.1&0.91\\    
C++:Dictionary&-2.1 (+/- 0.2)&-9.9&<\textbf{0.0001}\\
C\#:Dictionary&-2.42 (+/- 0.3)&-7.7&<\textbf{0.0001}\\
D:Dictionary&-13.22 (+/- 121.4)&-0.11&0.91\\    
Go:Dictionary&-4.51 (+/- 1.0)&-4.5&<\textbf{0.0001}\\
Java:Dictionary&-2.02 (+/- 0.2)&-8.3&<\textbf{0.0001}\\
Julia:Dictionary&-2.76 (+/- 0.4)&-6.6&<\textbf{0.0001}\\
JavaScript:Dictionary&-2.41 (+/- 0.2)&-10.6&<\textbf{0.0001}\\
Lua:Dictionary&0.88 (+/- 0.1)&8.9&<\textbf{0.0001}\\
Perl:Dictionary&-1.56 (+/- 0.2)&-7.1&<\textbf{0.0001}\\
PHP:Dictionary&0.28 (+/- 0.1)&2.7&\textbf{0.006}\\ 
R:Dictionary&-1.55 (+/- 0.3)&-4.7&<\textbf{0.0001}\\
Ruby:Dictionary&-0.18 (+/- 0.1)&-1.4&0.17\\    
Racket:Dictionary&-12.52 (+/- 112.5)&-0.11&0.91\\    
Rust:Dictionary&1.61 (+/- 0.1)&16.9&<\textbf{0.0001}\\
Scala:Dictionary&-1.24 (+/- 0.2)&-7.3&<\textbf{0.0001}\\
Swift:Dictionary&-0.19 (+/- 0.2)&-1.0&0.30\\    
TypeScript:Dictionary&-2.39 (+/- 0.2)&-10.049&<\textbf{0.0001}\\\hline
    \end{tabular}
    \caption{Mixed-effects results for the impact of language, continued features}\label{tab:mem_features_large2}
\end{table}

\section{Characterization of Code Generation Errors}\label{app:errors}

This section provides details regarding our error evaluation study on \newHE as overviewed in Section \ref{subsec:errors}. First we discuss the process of categorizing errors in a multi-language context. Then we provide the full set of themes, errors, and counts across the four studied languages: Python (\textsc{High}, untyped), C\# (\textsc{Medium}, typed), Swift (\textsc{Low}, typed), and Racket (\textsc{Niche}, untyped).  Finally, we showcase full code examples generated by Codex containing a variety of errors.

\subsection{Notes on Process \& Findings}


To perform the evaluation, we chose two typed languages and two untyped languages across all four frequency categories. A language expert then performed a manual investigation of a subset of the completions to derive a set of common error types. These errors could be associated with common error labels in a language (e.g., \texttt{NameError} in Python) or an observed phenomenon (e.g., \texttt{UseofDeprecatedIdentifiers} in Swift). Then, through an iterative process of manual inspection and automatic error detection via analyzing evaluation output, we developed a set of error labels unique to each language. We then arrived at the multi-language themes and categories via discussion and consensus.

The multi-language nature of the evaluation contributes to variation between the language classifications. For instance, languages vary significantly in the specificity of their error messages. Consider the theme of \texttt{TimeoutOrInfiniteRecursion}: Python has a specific error message \texttt{RecursionError} when it encounters an infinite recursive loop, whereas Racket will simply evaluate indefinitely. As the generated standard output and standard error were used for automatic classifications, there may be variations in how errors were counted depending on the error messages and precision of string search terms.

Overall, each error label is specific to the language under study and was subject to different levels of manual assessment. Therefore, the prevalence of a theme, rather than a specific error label or even category, likely provides a better source of inter- and intra-language information. Although the four languages in our study address different language variations (typed/untyped, frequency), they are not representative of all languages in our benchmark nor additional unstudied languages. Therefore, it is likely there are error labels, themes, and potentially categories that are missing from this characterization. Errors classified under the theme ``AssertionFailed'' describe errors from generated code with correct syntax which produces incorrect output. Other than via manual inspection of the over 10,000+ errors per language, there is no clear method of more precisely classifying errors of that type.

\subsection{Complete Error Themes}

In Tables \ref{tab:py_errors} - \ref{tab:rkt_errors} below, rows with the \colorbox{black!15}{gray} background are the most frequent error in that category for the specific language. Items in \emph{italics} are errors directly referenced in Section \ref{subsec:errors}. There are around 32,000 Codex completions for each language for our full translation. Variations in the reported counts below are due to support for a different number of prompts for each language and completions which generate multiple errors on failure. In the later case, we count all present errors.

\begin{center}
\begin{table}[H]
    \centering
    \begin{tabular}{|l|l|l|l|l|}
    \hline
    \textbf{Category}&\textbf{Theme}&\textbf{Error}&\textbf{Count}&\textbf{Example}\\\btrule{2pt}
    \rowcolor{black!15}
Runtime&AssertionFailed&AssertionError&17104&\\ \hline
Runtime&TimeoutOrInfiniteRecursion&Timeout&4462&\\ \hline
Runtime&InvalidDataStructureOperation&IndexError&460&\\ \hline
Runtime&TimeoutOrInfiniteRecursion&RecursionError&86&\\ \hline
Runtime&InvalidDataStructureOperation&AttributeError&5&\\ \hline
Runtime&InvalidDataStructureOperation&KeyError&2&\\ \hline
Runtime&DivisionByZero&ZeroDivisionError&1&\\ \btrule{2pt}
   \rowcolor{black!15}
Static&\emph{UndefinedIdentifier}&NameError&2942&Fig. \ref{fig:import-error}\\ \hline
Static&\emph{UndefinedIdentifier}&UnboundLocalError&1&\\ \btrule{2pt}
   \rowcolor{black!15}
Type&InvalidTypeConversion&TypeError&123&\\ \btrule{2pt}
   \rowcolor{black!15}
Language&Miscellaneous&ValueError&334&\\ \hline
Language&Miscellaneous&IndentationError&32&\\ \hline
Language&LanguageSpecific&EOFError&1&Fig. \ref{fig:python-eof}\\ \btrule{2pt}
   \rowcolor{black!15}
Model&OutOfTokens&SyntaxError&333&\\ \hline
Model&\emph{ExceptionInGeneratedCode}&NotImplementedError&253&Fig. \ref{fig:exception-purpose}\\ \hline
\end{tabular}
\caption{Error Categories, Themes, and Labels for Python }
\label{tab:py_errors}
\end{table}

\end{center}

\begin{center}
\begin{table}[H]
    \centering
    \begin{tabular}{|l|l|l|l|l|}
    \hline
    \textbf{Category}&\textbf{Theme}&\textbf{Error}&\textbf{Count}&\textbf{Example}\\\btrule{2pt}
     \rowcolor{black!15}
Runtime&AssertionFailed&AssertionError&22473&\\ \hline
Runtime&TimeoutOrInfiniteRecursion&Timeout&4470&\\ \hline
Runtime&NullReference&NullReferenceException&1201&\\ \hline
Runtime&InvalidDataStructureOperation&ArgumentOutOfRangeException&632&\\ \hline
Runtime&InvalidDataStructureOperation&InvalidOperationException&93&\\ \hline
Runtime&InvalidDataStructureOperation&IndexOutOfRangeException&82&\\ \hline
Runtime&InvalidDataStructureOperation&KeyNotFoundException&4&\\ \btrule{2pt}
 \rowcolor{black!15}
Static&\emph{UndefinedIdentifier}&UndefinedIdentifier&1577&\emph{Fig. \ref{fig:local-context}}\\ \hline
Static&MissingReturn&MissingReturn&155&\\ \hline
Static&UndefinedIdentifier&MethodNotFound&40&\\ \hline
Static&UndefinedIdentifier&TypeNotFound&15&\\ \hline
Static&ArityMismatch&InvalidArgument&11&\\ \hline
Static&ReDeclaration&ReDeclaration&2&\\ \btrule{2pt}
 \rowcolor{black!15}
Type&InvalidTypeConversion&TypeConversion&409&\\ \btrule{2pt}
 \rowcolor{black!15}
Language&Miscellaneous&FormatException&77&\\ \hline
Language&LanguageSpecific&InvalidAssignment&13&\\ \hline
Language&Miscellaneous&ArgumentException&1&\\ \btrule{2pt}
 \rowcolor{black!15}
Model&OutOfTokens&SyntaxError&319&\\ \hline
Model&\emph{ExceptionInGeneratedCode}&NotImplementedException&5&\\ \hline
Model&\emph{ExceptionInGeneratedCode}&InvalidBeat&1&\\ \hline
   \end{tabular}
    \caption{Error Categories, Themes, and Labels for C\# }
    \label{tab:csharp_errors}
\end{table}

\end{center}

\begin{center}
\begin{table}[H]
    \centering
    \begin{tabular}{|l|l|l|l|l|}
    \hline
    \textbf{Category}&\textbf{Theme}&\textbf{Error}&\textbf{Count}&\textbf{Example}\\\btrule{2pt}
     \rowcolor{black!15}
Runtime&AssertionFailed&AssertionFail&10051&     \\ \hline
Runtime&InvalidDataStructureOperation&IndexOutOfRange&330&\\ \hline
Runtime&TimeoutOrInfiniteRecursion&Timeout&275&\\ \hline
Runtime&InvalidDataStructureOperation&InvalidRangeCreation&271&\\ \hline
Runtime&NullReference&UnwrapNil&149&\\ \hline
Runtime&InvalidDataStructureOperation&StringIndexOutOfBounds&99&\\ \hline
Runtime&DivisionByZero&DivisionByZeroInRemainder&24&\\ \hline
Runtime&InvalidDataStructureOperation&RemoveLastFromEmptyCollection&6&\\ \hline
Runtime&InvalidDataStructureOperation&ArrayIndexOutOfRange&3&\\ \hline
Runtime&InvalidDataStructureOperation&RemoveFirstFromEmptyCollection&1&\\ \hline
Runtime&InvalidDataStructureOperation&NegativeArrayIndex&1&\\\btrule{2pt}
\rowcolor{black!15}
Static&\emph{UndefinedIdentifier}&CanNotFindInScope&4259&\\ \hline
Static&\emph{UndefinedIdentifier}&NonExistentMethod&2582&\\ \hline
Static&\emph{UndefinedIdentifier}&InvalidSyntax&213&\\ \hline
Static&\emph{UndefinedIdentifier}&CallingNonFunctionType&103&\\ \hline
Static&IncorrectAPIMethodCall&SubscriptStringWithInt&68&Fig. \ref{fig:bad-swift}\\\hline
Static&\emph{UndefinedIdentifier}&LinkerError&55&\\ \hline
Static&IncorrectAPIMethodCall&StringsArentCharArrays&42&\\ \hline
Static&\emph{UndefinedIdentifier}&UseBeforeDecl&17&\\ \hline
Static&IncorrectAPIMethodCall&StringIndices&12&\\ \hline
Static&ReDeclaration&RedeclarationOfVariable&11&\\\btrule{2pt}
\rowcolor{black!15}
Type&InvalidTypeConversion&OtherLocation&556&\\ \hline
Type&InvalidTypeConversion&ReturnTypeError&349&\\ \hline
Type&InvalidTypeConversion&ArgumentTypeError&303&\\ \hline
Type&InvalidTypeConversion&NumericsTypeError&261&\\ \hline
Type&InvalidTypeConversion&CollectionAndInner&241&\\ \hline
Type&InvalidTypeConversion&UnknownTypeErrorInCall&200&\\ \hline
Type&InvalidTypeConversion&BinOpTypeError&182&\\ \hline
Type&InvalidTypeConversion&BranchTypeMismatch&125&\\ \hline
Type&InvalidTypeConversion&MiscTypeError&111&\\ \hline
Type&InvalidTypeConversion&UnwrappedNonOptional&67&\\ \hline
Type&InvalidTypeConversion&UseOfModWithFloat&63&\\ \hline
Type&InvalidTypeConversion&ClosureResultTypeError&11&\\ \hline
Type&InvalidTypeConversion&ShouldHaveUnwrappedOptional&10&Fig. \ref{fig:type-system}\\\hline
Type&InvalidTypeConversion&PatternTypeError&10&\\ \hline
Type&InvalidTypeConversion&AssignmentTypeError&10&\\ \hline
Type&InvalidTypeConversion&WeirdSubscriptTypeError&10&\\ \hline
Type&InvalidTypeConversion&SubscriptingTypeError&9&\\\btrule{2pt}
\rowcolor{black!15}
Language&LanguageSpecific&UseOfDeprecatedIdentifiers&176&\\ \hline
Language&\emph{LanguageSpecific}&\emph{MissingArgumentLabel}&113&\emph{Fig. \ref{fig:label-swift}}\\\hline
Language&LanguageSpecific&ImmutableViolation&62&\\ \hline
Language&LanguageSpecific&ExtraArgument&38&\\ \hline
Language&LanguageSpecific&IncorrectArgumentLabel&19&\\ \hline
Language&LanguageSpecific&OverflowUnderflowTrap&9&\\ \hline
Language&LanguageSpecific&ExtraneousArgumentLabel&5&\\ \hline
Language&LanguageSpecific&NonExclusiveMutation&3&\\\btrule{2pt}
\rowcolor{black!15}
Model&OutOfTokens&RanOutOfTokens&95&\\ \hline
Model&OutOfTokens&CompilerErrorCutoff&9&\\ \hline
Model&OutOfTokens&MissingReturn&4&\\ \hline

    \end{tabular}
    \caption{Error Categories, Themes, and Labels for Swift }
    \label{tab:swift_errors}
\end{table}

\end{center}

\begin{center}
\begin{table}[H]
    \centering
    \begin{tabular}{|l|l|l|l|l|}
    \hline
    \textbf{Category}&\textbf{Theme}&\textbf{Error}&\textbf{Count}&\textbf{Example}\\\btrule{2pt}
    \rowcolor{black!15}
Runtime&AssertionFailed&assertionError&10409&\\ \hline
Runtime&TimeoutOrInfiniteRecursion&timeout&1044&\\ \hline
Runtime&InvalidDataStructureOperation&stringIndexOutOfRange&448&\\ \hline
Runtime&DivisionByZero&divisionBy0&376&\\ \hline
Runtime&InvalidDataStructureOperation&letDuplicateIdentifier&2&\\ \btrule{2pt}
 \rowcolor{black!15}
Static&\emph{UndefinedIdentifier}&unboundIdentifier&5814& \emph{Fig. \ref{fig:pythonic}}\\ \hline
Static&ArityMismatch&arityMismatch&570&\\ \hline
Static&ReDeclaration&listIndexOutOfRange&281&\\ \hline
Static&\emph{UndefinedIdentifier}&applicationOnNotAFunction&184&\\ \hline
Static&IncorrectAPIMethodCall&badBuiltInFnAppCall&69&\\ \hline
Static&\emph{UndefinedIdentifier}&useBeforeInit&12&\\ \btrule{2pt}
 \rowcolor{black!15}
Type&InvalidTypeConversion&contractViolation&4887&\\ \hline
Type&InvalidTypeConversion&hashmapNoValueForKey&7&\\ \btrule{2pt}
 \rowcolor{black!15}
Language&\emph{DoesNotKnowSyntax}&missingOrUnmatchedBraces&2004&\\ \hline
Language&LanguageSpecific&unknownEscapeSequence&216&\\ \hline
Language&\emph{DoesNotKnowSyntax}&badSyntax&127&\\ \hline
Language&\emph{DoesNotKnowSyntax}&missingElse&63&\\ \hline
Language&\emph{DoesNotKnowSyntax}&ifExtraArgs&48&\\ \hline
Language&Miscellaneous&badMatch&24&\\ \hline
Language&\emph{DoesNotKnowSyntax}&unclassified&6&\\ \hline
Language&LanguageSpecific&badCharConst&4&\\ \hline
Language&\emph{DoesNotKnowSyntax}&elseAsExpression&2&\\ \btrule{2pt}
 \rowcolor{black!15}
Model&OutOfTokens&incompleteGeneration&2092&\\ \hline
Model&\emph{GenerateAnotherLang}&generatingWebpageOrMarkdown&608&Fig. \ref{fig:racket-lang}\\ \hline
Model&GenerateAnotherLang&defaultValueExpMissing&57&\\ \hline
Model&ExceptionInGeneratedCode&errorByGeneratedProgram&34&\\\hline
\end{tabular}
\caption{Error Categories, Themes, and Labels for Racket }
\label{tab:rkt_errors}
\end{table}

\end{center}

\subsection{Error Examples Related to Section \ref{subsec:errors}}

\begin{figure}[H]
    \centering
\begin{lstlisting}[numbers=none]
class Problem {
    // Translated docstring
    public static string Intersection(Tuple<long, long> interval1, Tuple<long, long> interval2) {
        long start = Math.Max(interval1.Item1, interval2.Item1);
        long end = Math.Min(interval1.Item2, interval2.Item2);
        if (start > end) {
            return "NO";
        }
        long length = end - start + 1;
        return IsPrime(length) ? "YES" : "NO";
    }
\end{lstlisting}
    \caption{\emph{C\#}: Exhibits a local context error as it relies on a non-existent \texttt{IsPrime} method (HumanEval Prompt \#127, Theme: ``UndefinedIdentifier'', Code: ``UndefinedIdentifier'')}
    \label{fig:local-context}
\end{figure}

\begin{figure}[H]
    \centering
\begin{lstlisting}[numbers=none]
#lang racket

;; You are given a non-empty list of positive integers. Return the 
greatest integer that is greater than 
;; zero, and has a frequency greater than or equal to the value of the
integer itself. 
;; The frequency of an integer is the number of times it appears in the list.
;; If no such a value exist, return -1.
;; Examples:
;; >>> (search (list 4 1 2 2 3 1))
;; 2
;; >>> (search (list 1 2 2 3 3 3 4 4 4))
;; 3
;; >>> (search (list 5 5 4 4 4))
;; -1
(define (search lst)
  (let ([freq (make-hash)])
    (for ([i lst])
      (hash-set! freq i (+ 1 (hash-ref freq i 0))))
    (let ([max-freq (apply max (hash-values freq))])
      (for ([i lst])
        (when (and (> i 0) (= max-freq (hash-ref freq i)))
          (return i))))
    -1))

\end{lstlisting}
    \caption{\emph{Racket}: A completion which uses the keyword \texttt{return}. \texttt{return} is not a keyword in Racket, but is a common keyword in Python. (HumanEval Prompt \#69, Theme: ``UndefinedIdentifier'', Error: ``unboundIdentifier'' )}
    \label{fig:pythonic}
\end{figure}

\begin{figure}[H]
    \centering
\begin{lstlisting}[numbers=none]
def same_chars(s0: str, s1: str) -> bool:
    """
    docstring
    """
    return Counter(s0) == Counter(s1)
\end{lstlisting}
    \caption{\emph{Python}: This code expects a call to \texttt{import collections} to facilitate the use of \texttt{Counter} dictionary subclass (HumanEval Prompt \# 54, Theme: ``UndefinedIdentifier'', Error: ``NameError'') }
    \label{fig:import-error}
\end{figure}

\begin{figure}[H]
    \centering
\begin{lstlisting}[numbers=none]
def find_closest_elements(numbers: List[float]) -> Tuple[float, float]:
    """ From a supplied list of numbers (of length at least two) select and return two
    that are the closest to each other and return them in 
    order (smaller number, larger number).
    >>> find_closest_elements([1.0, 2.0, 3.0, 4.0, 5.0, 2.2])
    (2.0, 2.2)
    >>> find_closest_elements([1.0, 2.0, 3.0, 4.0, 5.0, 2.0])
    (2.0, 2.0)
    """
    # YOUR CODE HERE
    raise NotImplementedError()
\end{lstlisting}
    \caption{\emph{Python}: The full completion for this program contains only a call to \texttt{raise}, so this code will always produce an error. (HumanEval Prompt \#20, Theme: ``ExceptionInGeneratedCode'', Error: ``NotImplementedError'') }
    \label{fig:exception-purpose}
\end{figure}

\begin{figure}[H]
    \centering
\begin{lstlisting}[numbers=none]
#lang racket 

;; You are given a string s.
;; if s[i] is a letter, reverse its case from lower to upper or vise versa, 
;; otherwise keep it as it is.
;; If the string contains no letters, reverse the string.
;; The function should return the resulted string.
;; Examples
;; >>> (solve "1234")
;; "4321"
;; >>> (solve "ab")
;; "AB"
;; >>> (solve "#a@C")
;; "#A@c"
(define (solve s)
  (if (string-contains s #\a)
      (string-map (lambda (c) (if (char-alphabetic? c)
                                  (char-upcase c)
                                  (char-downcase c)))
                  s)
      (string-reverse s)))
```

### [Reverse the order of words in a string]
(https://www.codewars.com/kata/reverse-the-order-of-words-in-a-string)

```racket
#lang racket
\end{lstlisting}
    \caption{\emph{Racket}: This snippet is an example of Markdown being generated instead of Racket (it is word wrapped for presentation here). (HumanEval Prompt \#161, Theme: ``GenerateAnotherLanguage'', Error: ``GeneratingWebpageOrMarkdown'')}
    \label{fig:racket-lang}
\end{figure}

\begin{figure}[H]
    \centering
\begin{lstlisting}[numbers=none]
/// This function takes an array l and returns an array l' such that
/// l' is identical to l in the indicies that are not divisible by three, 
while its values at the indicies that are divisible by three are equal
/// to the values of the corresponding indicies of l, but sorted.
/// >>> sort_third(l: [1, 2, 3])
/// [1, 2, 3]
/// >>> sort_third(l: [5, 6, 3, 4, 8, 9, 2])
/// [2, 6, 3, 4, 8, 9, 5]
func sort_third(l: [Int]) -> [Int] {
    var l = l
    for i in stride(from: 2, to: l.count, by: 3) {
        l[i] = l[i].min(l[i - 1])
        l[i - 1] = l[i].max(l[i - 1])
    }
    return l
}
\end{lstlisting}
    \caption{\emph{Swift}: This code produces the type error \texttt{error: value of optional type 'Int?' must be unwrapped to a value of type 'Int'}
    (HumanEval Prompt \# 33, Theme: ``InvalidTypeConversion'', Error:``ShouldHaveUnwrappedOptional'')}
    \label{fig:type-system}
\end{figure}

\begin{figure}[H]
    \centering
\begin{lstlisting}[numbers=none]
/// Given a positive integer n, return the count of the numbers of n-digit
/// positive integers that start or end with 1.
func starts_one_ends(n: Int) -> Int {
    if n == 1 {
        return 2
    }
    if n == 2 {
        return 4
    }
    return starts_one_ends(n - 1) + starts_one_ends(n - 2)
}
\end{lstlisting}
    \caption{\emph{Swift}: A correct recursive call to \texttt{starts\_one\_ends} in Swift would be \texttt{starts\_one\_ends(n: n - 1)}. (HumanEval Prompt \#83, Theme: ``LanguageSpecific'', Error: ``MissingArgumentLabel'')}
    \label{fig:label-swift}
\end{figure}

\subsection{Additional Error Examples from our Codex Error Evaluation}

\begin{figure}[H]
    \centering
\begin{lstlisting}[numbers=none]
/// Translated docstring
func match_parens(lst: [String]) -> String {
    var lst = lst
    var str = ""
    var count = 0
    for i in 0..<lst.count {
        str += lst[i]
    }
    for i in 0..<str.count {
        if str[i] == "(" {
            count += 1
        } else {
            count -= 1
        }
        if count < 0 {
            return "No"
        }
    }
    if count == 0 {
        return "Yes"
    } else {
        return "No"
    }
}

\end{lstlisting}
    \caption{\emph{Swift}: This is an example of incorrect string indexing in Swift - you cannot call \texttt{str[i]} where \texttt{i} is an \texttt{Integer} (HumanEval Prompt \#119, Theme: ``IncorrectAPIMethodCall'', Error: ``SubscriptStringWithInt'')}
    \label{fig:bad-swift}
\end{figure}

\begin{figure}[H]
    \centering
\begin{lstlisting}
def car_race_collision(n: int) -> int:
    """
    Imagine a road that's a perfectly straight infinitely long line.
    n cars are driving left to right;  simultaneously, a different set of n cars
    are driving right to left.   The two sets of cars start out being very far from
    each other.  All cars move in the same speed.  Two cars are said to collide
    when a car that's moving left to right hits a car that's moving right to left.
    However, the cars are infinitely sturdy and strong; as a result, they continue moving
    in their trajectory as if they did not collide.

    This function outputs the number of such collisions.
    """
    # The idea is to sort the cars by their starting positions.
    # Then, we can simulate the race by iterating through the cars
    # from left to right, and counting the number of cars that are
    # moving right to left.
    #
    # The time complexity is O(n log n) because of the sorting.
    # The space complexity is O(n) because of the sorting.
    #
    # The time complexity can be improved to O(n) by using a
    # counting sort.  The space complexity can be improved to
    # O(1) by using a bucket sort.
    left_to_right = [0] * n
    right_to_left = [0] * n
    for i in range(n):
        left_to_right[i] = int(input())
        right_to_left[i] = int(input())
    left_to_right.sort()
    right_to_left.sort()
    collisions = 0
    left_to_right_index = 0
    right_to_left_index = 0
    while left_to_right_index < n and right_to_left_index < n:
        if left_to_right[left_to_right_index] < right_to_left[right_to_left_index]:
            collisions += right_to_left_index
            left_to_right_index += 1
        else:
            right_to_left_index 
\end{lstlisting}
    \caption{\emph{Python}: This code produces an \texttt{EOFError} due to the calls to \texttt{input()} inside the \texttt{for} loop. The canonical example does not use \texttt{input()}.  Our evaluation procedures do not handle user input, hence the error. (HumanEval Prompt \#41, Theme: ``LanguageSpecific'', Error: ``EOFError'') }
    \label{fig:python-eof}
\end{figure}


\lstdefinestyle{customStyle}{
basicstyle=\footnotesize\ttfamily,
keywordstyle=\bfseries,
showstringspaces=false,
morekeywords={include, printf},
commentstyle={},
escapeinside=QQ,
escapebegin=\begin{otherlanguage*}{bulgarian}\normalfont,
escapeend=\end{otherlanguage*},
}

\lstset{style=customStyle}
\begin{figure}[H]
    \centering
\begin{lstlisting}
;; You are given a list of integers.
;; You need to find the largest prime value and return the sum of its digits.
;; Examples:
;; >>> (skjkasdkd (list 0 3 2 1 3 5 7 4 5 5 5 2 181 32 4 32 3 2 32 324 4 3))
;; 10
;; >>> (skjkasdkd (list 1 0 1 8 2 4597 2 1 3 40 1 2 1 2 4 2 5 1))
;; 25
;; >>> (skjkasdkd (list 1 3 1 32 5107 34 83278 109 163 23 2323 32 30 1 9 3))
;; 13
;; >>> (skjkasdkd (list 0 724 32 71 99 32 6 0 5 91 83 0 5 6))
;; 11
;; >>> (skjkasdkd (list 0 81 12 3 1 21))
;; 3
;; >>> (skjkasdkd (list 0 8 1 2 1 7))
;; 7
(define (skjkasdkd lst)
  (define (is-prime? n)
    (define (iter i)
      (cond ((= i n) #t)
            ((= (modulo n i) 0) #f)
            (else (iter (+ i 1)))))
    (iter 2))
  (define (sum-digits n)
    (define (iter n acc)
      (cond ((= n 0) acc)
            (else (iter (quotient n 10) (+ acc (remainder n 10))))))
    (iter n 0))
  (define (largest-prime lst)
    (cond ((null? lst) 0)
          ((is-prime? (car lst)) (max (car lst) (largest-prime (cdr lst))))
          (else (largest-prime (cdr lst)))))
 (sum-digits (largest-prime lst)))
```
QQ
## QЗадачаQ 2

QДа се дефинира функцияQ `(skjkasdkd lst)`, Qкоято приема списък от цели числа и връща списък от цели числа, които са прости и по-големи от най-голямото просто число в дадения списък.Q

QПримери:Q
QQ

```racket
>>> (skjkasdkd (list 0 3 2 1 3 5 7 4 5 5 5 2 181 32 4 32 3 2 32 324 4 3))
'(181 191 193 197 199 211 223 227 229 233 239 241 251 257 263 
269 271 277 281 283 293 307 311 313 317 331 337 347 349 353 
359 367 373 379 383 389 397 401 409 419 421 431 433

\end{lstlisting}
    \caption{\emph{Racket}: Codex can produce comments/code in languages other than English. The Bulgarian text above translates as the following via Google Translate: ``\#\# Task 2, Define a function `(skjkasdkd lst)` that accepts a list of integers and returns a list of integers that are prime and greater than the largest prime in the given list., Examples:'' (HumanEval Prompt \#119, Theme: ``GenerateAnotherLanguage'', Error: ``GeneratingWebpageOrMarkdown'')}
    \label{fig:bad-rkt}
\end{figure}

\subsection{Additional Error Examples}

This section contains error examples which we found through manual inspection of Codex output outside of our structured analysis.

\lstdefinelanguage{JavaScript}{
sensitive=false,
morekeywords={let,function,return,undefined,null,typeof,true,false,ref,deref,
              var,new,break,while,try,catch,finally,throw,err,if,else,for,do,
              with,delete,new,instanceof,this,},
keywordstyle=\ttfamily,
identifierstyle=\ttfamily,
comment=[l]{//},
commentstyle=\ttfamily,
string=[d]{"},
stringstyle=\ttfamily,
mathescape=true,
extendedchars=true,
basicstyle=\footnotesize\ttfamily,
showstringspaces=false,
numbers=none,
firstnumber=0,
numberstyle=\tiny,
stepnumber=5,
numbersep=5pt,
upquote=true,
columns=fixed,
flexiblecolumns=true,
}

\begin{figure}[H]
\centering
\begin{lstlisting}[language=JavaScript,numbers=none]
function sum_squares(lst){
    return sum(map(lambda n: n ** 2, filter(lambda n: n % 3 == 0, lst)))
}
\end{lstlisting}
    \caption{This JavaScript completion is generated for our full translation at $T=0.8$ for HumanEval Prompt \#142. The complete function body is executable, functional-style Python.}
    \label{fig:javascript-python}
\end{figure}

\begin{figure}[H]
\centering
\begin{lstlisting}[language=JavaScript,numbers=none]
local function will_it_fly(q, w)
   return is_balanced(q) and sum(q) <= w
end
\end{lstlisting}
    \caption{This Lua completion is generated by our full translation at $T=0.2$ for HumanEval Prompt \#72. \texttt{is\_balanced} is not defined in the local context and is likely an expected helper function. The call to \texttt{sum} is Python-like, as \texttt{sum} is a Python built-in method.}
    \label{fig:lua-python}
\end{figure}

\end{document}